\documentclass[12pt]{article} %

\usepackage{times}
\usepackage{url}            %
\usepackage{booktabs}       %
\usepackage{amsfonts}       %
\usepackage{nicefrac}       %
\usepackage{microtype}      %
\usepackage{mkolar_definitions}
\usepackage{xcolor}
\usepackage{xspace}
\usepackage{multirow}
\usepackage{amsmath}
\usepackage{algorithm}
\usepackage{algorithmic}
\usepackage{color}
\usepackage{authblk}
\usepackage{color, colortbl}
\usepackage{enumitem}
\usepackage{comment}
\usepackage{bm}
\usepackage{wrapfig}
\usepackage{subcaption}

\newcommand{\spp}{\mathrm{sp}}
\newcommand{\KL}{\mathrm{KL}}

\newcommand{\alg}{BRAID}
\newcommand{\un}{\mathrm{un}}
\newcommand{\pre}{\mathrm{pre}}

\definecolor{cerise}{rgb}{0.87, 0.19, 0.39}
\definecolor{babyblue}{rgb}{0.54, 0.81, 0.94}
\definecolor{ao(english)}{rgb}{0.0, 0.5, 0.0}
\usepackage{enumitem}
\setlist[itemize]{leftmargin=5.5mm}

\usepackage{microtype}
\usepackage{graphicx}
\usepackage{booktabs} %

\usepackage{hyperref}

\usepackage{amsmath}
\usepackage{amssymb}
\usepackage{mathtools}
\usepackage{amsthm}

\usepackage[capitalize,noabbrev]{cleveref}

\usepackage{natbib}
\usepackage[scaled=.9]{helvet}
\usepackage[left=2cm,top=2cm,right=2cm]{geometry}
\usepackage[textsize=tiny]{todonotes}

\usepackage{authblk}

\title{Bridging Model-Based Optimization and Generative Modeling via Conservative Fine-Tuning of Diffusion Models }

\author[1]{Masatoshi Uehara \thanks{Equal contribution:  uehara.masatoshi@gene.com,\,yz6292@princeton.edu  } }
\author[2]{Yulai Zhao$\,^{*}$}
\author[1]{Ehsan Hajiramezanali}
\author[1]{Gabriele Scalia}  
\author[1]{\\ Gökcen Eraslan} 
\author[1]{Avantika Lal }
\author[3]{Sergey Levine \thanks{Corresponding authors: sergey.levine@berkeley.edu, biancalt@gene.com  }}
\author[1]{Tommaso Biancalani$\,^{\dagger}$  } 
\affil[1]{Genentech}
\affil[2]{Princeton University   }
\affil[3]{University of California Berkeley   }

\begin{document}

\maketitle

\begin{abstract}
AI-driven design problems, such as DNA/protein sequence design, are commonly tackled from two angles: generative modeling, which efficiently captures the feasible design space (e.g., natural images or biological sequences), and model-based optimization, which utilizes reward models for extrapolation. To combine the strengths of both approaches, we adopt a hybrid method that fine-tunes cutting-edge diffusion models by optimizing reward models through RL. Although prior work has explored similar avenues, they primarily focus on scenarios where accurate reward models are accessible. In contrast, we concentrate on an offline setting where a reward model is unknown, and we must learn from static offline datasets, a common scenario in scientific domains. In offline scenarios, existing approaches tend to suffer from overoptimization, as they may be misled by the reward model in out-of-distribution regions. To address this, we introduce a conservative fine-tuning approach, BRAID, by optimizing a conservative reward model, which includes additional penalization outside of offline data distributions. Through empirical and theoretical analysis, we demonstrate the capability of our approach to outperform the best designs in offline data, leveraging the extrapolation capabilities of reward models while avoiding the generation of invalid designs through pre-trained diffusion models. 

\end{abstract}

\section{Introduction}

\label{sec:intro}
 Computational design involves synthesizing designs that optimize a particular reward function. This approach finds applications in various scientific domains, including DNA/RNA/protein design \citep{sample2019human,gosai2023machine,wu2024protein}. While physical simulations are often used in design problems, lacking extensive knowledge of underlying physical processes necessitates solutions that solely rely on experimental data. In these scenarios, we need an algorithm that synthesizes an improved design by utilizing a dataset of past experiments (i.e., \emph{static offline dataset}). Existing research has addressed computational design from two primary angles. The first angle is generative modeling such as diffusion models \citep{ho2020denoising}, which aim to directly model the distribution of valid designs by emulating the offline data. This approach allows us to model the space of ``valid'' designs (e.g., natural images, natural DNA sequences, foldable protein sequences \citep{avdeyev2023dirichlet}). The second angle is offline model-based optimization (MBO), which entails learning the reward model from static offline data and optimizing it with respect to design inputs \citep{brookes2019conditioning,trabucco2021conservative,angermueller2019model,linder2021fast,fannjiang2020autofocused,chen2022bidirectional}. This class of methods potentially enables us to surpass the best design observed in the offline data by harnessing the extrapolative capabilities of reward models.

In our work, we explore how the generative modeling and MBO perspectives could be reconciled, inspired by recent work on RL-based fine-tuning of diffusion models (e.g., \citet{black2023training,fan2023dpok}), which aims to finetune diffusion models by optimizing down-stream reward functions. Although these studies do not originally address computational design, we can potentially leverage the strengths of both perspectives. However, these existing studies often focus on scenarios where online reward feedback can be queried or accurate reward functions are available. Such approaches are not well-suited for the typical offline setting, where we lack access to true reward functions and need to rely solely on static offline data \citep{levine2020offline,kidambi2020morel,yu2020mopo}. In scientific fields, this offline scenario is common due to the high cost of acquiring feedback data. In such contexts, existing works for fine-tuning diffusion models may easily lead to overoptimization, where optimized designs are misled by the trained reward model from the offline data, resulting in out-of-distribution adversarial designs instead of genuinely high-quality designs.

 \begin{figure}[!t]
\centering 
\begin{subfigure}{.35\textwidth}
    \includegraphics[width=1.0\linewidth]{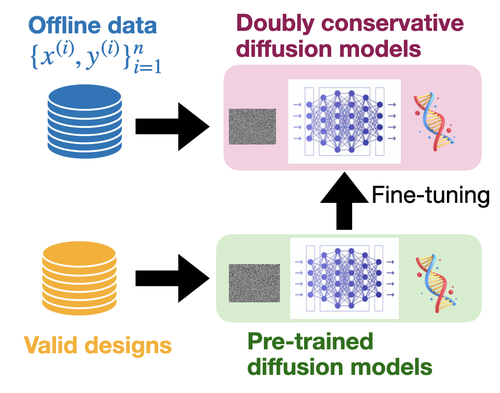} 
    \label{fig:setting}
    \end{subfigure}
    \begin{subfigure}{.45\textwidth}
    \centering
    \includegraphics[width=1.0\linewidth]{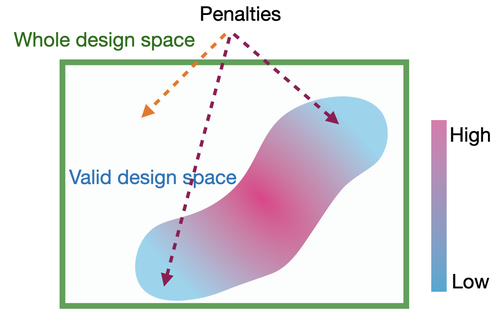} 
    \label{fig:algorithm}
    \end{subfigure}
 \caption{ { The left figure illustrates our setup with a pre-trained generative model and offline data. On the right, the motivation of the algorithm is depicted. The region surrounded by the \textcolor{ao(english)}{green} line is the original entire design space, with the colored region indicating the valid design space (e.g., natural images, human-like DNA sequences). The \textcolor{cerise}{red} region denotes areas with more offline data available, while the \textcolor{babyblue}{blue} region indicates areas with less data available. We aim to add penalties to the \textcolor{babyblue}{blue} regions using \textcolor{purple}{conservative reward modeling} to prevent overoptimization while imposing a \textcolor{orange}{stricter KL penalty} on the non-colored regions to prevent the generation of invalid designs.} 
 }
     \label{fig:algorithm}
\end{figure}

To mitigate overoptimization, we develop a conservative fine-tuning approach for generate models aimed at computational design.
Specifically, we consider a critical scenario where we have offline data (with feedback) and a pre-trained diffusion model capable of capturing the space of ``valid'' designs, and propose a two-stage method (Figure~\ref{fig:algorithm}).
In the initial stage, we train a conservative reward model using offline data, incorporating an uncertainty quantification term that assigns higher penalties to out-of-distribution regions. Subsequently, we finetune pre-trained diffusion models by optimizing the conservative reward model to obtain high-quality designs and prevent the generation of out-of-distribution designs. In the fine-tuning process, we also introduce a KL penalization term to ensure that the generated designs remain within the valid design space.

Our primary contribution lies in the introduction of a novel framework,  \textbf{\alg} (douBly conseRvAtive fine-tuning diffusIon moDels). The term ``doubly conservative'' reflects the incorporation of two types of conservative terms, both in reward modeling and KL penalization. By properly penalizing the fine-tuned diffusion model when it deviates significantly from the offline data distribution, we effectively address overoptimization. Additionally, by framing our fine-tuning procedure within the context of soft-entropy regularized Markov Decision Processes, we offer theoretical justification for the inclusion of these conservative terms in terms of regret.
This theoretical result shows that fine-tuned generative models outperform the best designs in the offline data, leveraging the extrapolation capabilities of reward models while avoiding the generation of invalid designs. Furthermore, through empirical evaluations, we showcase the efficacy of our approach across diverse domains, such as DNA/RNA sequences and images.

\section{Related Works}

We summarize related works. For additional works such as fine-tuning on LLMs, refer to Section~\ref{sec:additional}. 

\vspace{-3mm}
\paragraph{Fine-tuning diffusion models via reward functions.} Several previous studies have aimed to improve diffusion models by optimizing reward functions using various methods, including supervised learning \citep{lee2023aligning, wu2023better}, RL \citep{black2023training, fan2023dpok} and control-based techniques \citep{clark2023directly, xu2023imagereward,prabhudesai2023aligning}. In contrast to our work, their emphasis is not on an offline setting, i.e., their setting assumes online reward feedback is available or accurate reward functions are known. Additionally, while \citet{fan2023dpok} include the KL term in their algorithms, our innovation lies in integrating conservative reward modeling to mitigate overoptimization and formal statistical guarantees in terms of regret (Theorem~\ref{thm:key},\,\ref{thm:regret}).

\vspace{-2mm}
\paragraph{Conditional diffusion models.} Conditional diffusion models, which learn conditional distributions of designs given the rewards, have been extensively studied  \citep{ho2022classifier,dhariwal2021diffusion,song2020denoising,bansal2023universal}. However, for the purpose of MBO, these approaches require that the offline data has good coverage on values we want to condition on \citep{brandfonbrener2022does}. Compared to conditional diffusion models, our approach aims to obtain designs that can surpass the best design in offline data by leveraging the extrapolation capabilities of reward models. We compare these approaches with our work in Section \ref{sec:main_experiments}.

\vspace{-2mm}
\paragraph{Offline model-based optimization (MBO).} 

Offline MBO is also known as offline black-box optimization and is closely related to offline contextual bandits and offline RL \citep{levine2020offline}. While conservative approaches have been studied there (e.g., \citet{kidambi2020morel,yu2020mopo} and more in \pref{sec:additional}); most of the works are not designed to incorporate a diffusion model, unlike our approach. Hence, it remains unclear how these methods can generate designs that remain within intricate valid design spaces (e.g., generating natural images). 

It is worth noting a few exceptions \citep{yuan2023reward,krishnamoorthy2023diffusion} that attempt to integrate diffusion models into MBO. However, the crucial distinctions lie in the fact that we directly optimize rewards with diffusion models, whereas these prior works focus on using conditional diffusion models. 
Additionally, we delve into the incorporation of conservative terms, an aspect not explored in their works. We compare these methods with ours empirically in Section \ref{sec:main_experiments}.

\section{Preliminaries}

We outline our framework for offline model-based optimization with a pre-trained generative model. Subsequently, we highlight the challenges arising from distributional shift. Additionally, we provide an overview of diffusion models, as we will employ them as pre-trained generative models.

\subsection{Offline Model-Based Optimization with Pre-Trained Generative Model}
\label{subsec:pretrain}

Our objective is to find a high-quality design within a design space, $\Xcal$. Each design $x \in \Xcal$ is associated with a reward, $r(x)$, where $r:\Xcal \to [0,1]$ is an unknown reward function. Then, our aim is to find a high-quality generative model $p\in \Delta(\Xcal)$, that yields a high $r(x)$. It is formulated as  
\begin{align}\label{eq:original_goal}\textstyle
    \argmax_{p \in  \Delta(\Xcal)}\EE_{x\sim p}[r(x)]. 
\end{align}

\paragraph{Avoiding invalid designs.} In MBO, the design space $\Xcal$ is typically huge. However, in practice, the valid design space denoted by $\Xcal_{\pre}$ is effectively contained within this extensive $\Xcal$ as a potentially lower-dimensional manifold. For instance, in biology, our focus often centers around discovering highly bioactive protein sequences. While the raw search space might encompass $|20|^B$ possibilities (where $B$ is the length), the actual design space corresponding to valid proteins is significantly more constrained. Consequently, our problem can be formulated as:
\begin{align}\label{eq:hard}
    \argmax_{p \in  \Delta(\Xcal_{\pre})}\EE_{x\sim p}[r(x)],  (\text{eqivaletnly,\,}\argmax_{p \in \Delta(\Xcal)} \EE_{x\sim p} [r(x) ] - {\EE_{x\sim p}[\mathrm{I}(x \notin \Xcal_{\pre} ) ]}).  
\end{align}
Note supposing that a reward $r(\cdot)$ is $0$ outside of $\Xcal_{\pre}$, this is actually still equivalent to \eqref{eq:original_goal}. 

\paragraph{Offline data with a pre-trained generative model.}
Based on the above motivation, we consider scenarios where we have an offline dataset $\Dcal_{\mathrm{off}}$, used for learning the reward function. More specifically, the dataset, $ \Dcal_{\mathrm{off}}= \{x^{(j)},y^{(j)}\}_{j=1}^{n_{\mathrm{off}}}$ contains pairs of designs $x \sim p_{\mathrm{off}}(\cdot)$ and their associated noisy reward feedbacks $y=r(x)+\epsilon$, where $\epsilon$ is noise.  

Compared to settings in many existing papers on MBO, we also assume access to a pre-trained generative model (diffusion model) trained on a large dataset comprising valid designs, in addition to the offline data $\Dcal_{\mathrm{off}}$. For example, in biology, this is expected to capture the valid design space $\Xcal_{\pre}$ such as human DNA sequences or physically feasible proteins \citep{avdeyev2023dirichlet,li2024discdiff,sarkar2024designing,stark2024dirichlet,campbell2024generative}. These pre-trained generative models are anticipated to be beneficial for narrowing down the raw search space $\Xcal$ to the design space $\Xcal_{\pre}$. In our work, denoting the distribution induced by the pre-trained model by $p_{\pre}$, we regard the support of $p_{\pre}$ as $\Xcal_{\pre}$.

\subsection{Challenge: Distributional Shift}
\label{subsec:naive}

To understand our challenges, let's first examine a simple approach for MBO with a pre-trained generative model. For instance, we can adapt methods from \citet{clark2023directly,prabhudesai2023aligning} to our scenario. This approach involves two steps. In the first step, we perform reward learning: $
    \hat r   = \argmin_{\tilde r \in \Fcal}\sum_{i=1}^{n_{\mathrm{off}}}  \{\tilde r(x^{(i)})-y^{(i)}\}^2, $
where $\Fcal$ represents a function class that includes mappings from $\Xcal$ to $[0,1]$, aiming to capture the true reward function $r(\cdot)$. Then, in the second step, we fine-tune a pre-trained diffusion model to optimize $\hat r$.

Despite its simplicity, this approach faces two types of distributional shifts. Firstly, the fine-tuned generative model might produce invalid designs outside of $\Xcal_{\pre}$. As discussed in Section~\ref{subsec:pretrain}, we aim to prevent this situation. Secondly, the fine-tuned generative model may over-optimize $\hat r$, exploiting uncertain regions of the learned model $\hat r$. Indeed, in regions not covered by offline data distribution $p_{\mathrm{off}}$, the learned reward $\hat r$ can easily have higher values, while the actual reward values in terms of $r$ might be lower due to the higher uncertainty. We aim to avoid situations where we are misled by out-to-distribution adversarial designs.

\subsection{Diffusion Models}\label{subsec:diffusion}

We present an overview of denoising diffusion probabilistic models (DDPM) \citep{song2020denoising,ho2020denoising,sohl2015deep}. Note while the original diffusion model was initially introduced in Euclidean spaces, it has since been extended to simplex spaces for biological sequences \citep{avdeyev2023dirichlet}, which we will use in \pref{sec:main_experiments}. In diffusion models, the goal is to develop a generative model that accurately emulates the data distribution from the dataset. Specifically, denoting the data distribution by $p_{\pre} \in \Delta(\Xcal)$, a DDPM aims to approximate using a parametric model structured as  $p(x_0;\theta)=\int p(x_{0:T};\theta)d x_{1:T}$, where $p(x_{0:T};\theta) =p_{T+1}(x_T;\theta)\prod_{t=T}^1 p_{t}(x_{t-1}|x_t;\theta)$. Here, each $p_t$ is considered as a policy, which is a mapping from a design space $\Xcal$ to a distribution over $\Xcal$. By optimizing the variational bound on the negative log-likelihood, we can obtain a set of policies $\{p_t\}_{t=T+1}^1$ such that $p(x_0;\theta)\approx p_{\pre}(x_0)$. For simplicity, in this work, assuming that pre-trained diffusion models are accurate, we denote the pre-trained policy as $\{p^{\mathrm{pre}}_t(\cdot|\cdot)\}_{t=T+1}^1$, and the generated distribution by the pre-trained diffusion model at $x_0$ by $p_{\pre}$. With slight abuse of notation, we often denote $p^{\mathrm{pre}}_{T+1}(\cdot)$ by $p^{\mathrm{pre}}_{T+1}(\cdot|\cdot)$

\section{Doubly Conservative Generative Models}

We've discussed how na\"ive approaches for computational design may yield invalid designs or over-optimize reward functions, with both challenges stemming from distributional shift. Our goal in this section is to develop doubly conservative generative models to mitigate this distributional shift.

\subsection{Avoiding Invalid Designs }

To avoid invalid designs, we begin by considering the following generative model:
\begin{align}\label{eq:goal_objective} 
      \frac{\exp(\hat r(\cdot)/\alpha)p_{\pre}(\cdot)}{\int \exp(\hat r(x)/\alpha)p_{\pre}(x)dx }\quad (:= \argmax_{p \in \Delta(\Xcal)} \EE_{x \sim p} [\hat r(x)]-\alpha {\KL(p \|p_{\pre} )}),
\end{align}
where $\KL(p \|p_{\pre} )=\EE_{x \sim p}[\log(p(x)/p_{\pre}(x))]$. 
In this formulation, the generative model is designed as an optimizer of a loss function composed of two parts: the first component encourages designs with high rewards, while the second component acts as a regularizer penalizing the generative model for generating invalid designs. This formulation is inspired by our initial objective in \eqref{eq:hard}, where we substitute an indicator function with $\log(p/p_{\pre})$. This regularizer takes $\infty$ when $p$ is not covered by $p_{\pre}$, and $\alpha$ governs the strength of the regularizer. 

\subsection{Avoiding Overoptimization }

Next, we address the issue of overoptimization. This occurs when we are fooled by the learned reward model in uncertain regions. Therefore, a natural approach is to penalize generative models when they produce designs in uncertain regions.

As a first step, let's consider having an uncertainty oracle $\hat g:\Xcal \to [0,1]$, which is a random variable of $\Dcal_{\mathrm{off}}$. This oracle is expected to quantify the uncertainty of the learned reward function $\hat r$.

\begin{assum}[Uncertanity oracle]\label{assum:crucial}
 With probability $1-\delta$, we have 
 \begin{align}\label{eq:calibrated}
   \forall x\in \Xcal_{\pre};  |\hat r(x)-r(x)|\leq \hat g(x) 
 \end{align}
\end{assum} 
 
These calibrated oracles are well-established when using a variety of models such as linear models, Gaussian processes, and neural networks. We will provide detailed examples of such calibrated oracles in Section~\ref{subsec:oracles}. Essentially, as long as the reward model is well-specified (i.e., there exists $\tilde r \in \Fcal$ such that $\forall x\in \Xcal_{\pre}:\tilde r(x)=r(x)$), we can create such a calibrated oracle.

\paragraph{Doubly Conservative Generative Models.} 

Utilizing the uncertainty oracle defined in Assumption~\ref{assum:crucial}, we present our proposal:
\begin{align}\label{eq:doubly_conservative,eq:goal_objective}
     \hat \pi_{\alpha}(\cdot)=  \frac{\exp\left ( (\hat r-\hat g)(\cdot)/\alpha \right)p_{\pre}(\cdot)}{\int \exp\left ( (\hat r - \hat g)(x)/\alpha \right)p_{\pre}(x)dx }\,\, (:= \argmax_{p \in \Delta(\Xcal)} \underbrace{\EE_{x \sim p} [(\hat r-\hat g)(x)]}_{\text{Penalized\,reward } }- \underbrace{\alpha \KL(p \|p_{\pre} )}_{\text{KL Penalty } }). 
\end{align}
Here, to combat overoptimization, we introduce an additional penalty term $\hat g(x)$. This penalty term is expected to prevent $\hat \pi_{\alpha}$ from venturing into regions with high uncertainty because it would take a higher value in such regions. We refer to $\hat \pi_{\alpha}$ as a doubly conservative generative model due to the incorporation of two conservative terms.

An attentive reader might question the necessity of simultaneously introducing two conservative terms. Specifically, the first natural question is whether KL penalties, intended to prevent invalid designs, can replace uncertainty-oracle-based penalties. However, this may not hold true because even if we can entirely avoid venturing outside of $\Xcal_{\pre}$ (support of $p_{\pre}$), we may still output designs on uncertain regions not covered by $p_{\mathrm{off}}$. The second question is whether uncertainty-oracle-based penalties can substitute KL penalties. While it is partly true in situations where the support of $p_{\mathrm{off}}$ is contained within that of $p_{\pre}$, uncertainty-oracle-based penalties, lacking leverage on pre-trained generative models, are ineffective in preventing invalid designs. In contrast, KL penalties are considered a more direct approach to stringently avoid invalid designs by leveraging pre-trained generative models.

\subsection{Examples of Uncertainty Oracles}\label{subsec:oracles}

\begin{example}[Gaussian processes.]\label{exa:gps}
When we use an RKHS as $\Fcal$ (a.k.a. GPs) associated with a kernel $k(\cdot,\cdot):\Xcal \times \Xcal \to \RR$ \citep{srinivas2009gaussian}, a typical construction of $\hat r$ and $\hat g$ is 
 \begin{align*}\textstyle 
     \hat r(\cdot)=\Yb (\Kb +\lambda I)^{-1}\kb(\cdot),\,\hat g(\cdot)=c(\delta)\sqrt{\hat k(\cdot,\cdot)}, 
 \end{align*}
 where $c(\delta) \in \RR_{>0},\lambda \in \RR_{>0}$, $\kb(x)=[k(x^{(1)},x),\cdots,k(x^{(n_{\mathrm{off}})},x)]^{\top}$, 
 \begin{align*}
   \Yb=[y^{(1)},\cdots , y^{(n_{\mathrm{off}})}],  \{ \Kb \}_{p,q} = k(x^{(p)},x^{(q)}), 
  \hat k(x,x')= k(x,x')-\kb(x)^{\top}\{ \Kb+\lambda I \}^{-1}  \kb(x'). 
 \end{align*}
Note that when using deep neural networks, by considering the last layer as a feature map, we can still create a kernel \citep{zhang2022making,qiu2022contrastive}. 
\end{example}

\begin{example}[Bootstrap]\label{exa:bootstrap}
When we use neural networks as $\Fcal$, it is common to use a statistical bootstrap method. Note many variants have been proposed \citep{chua2018deep}, and its theory has been analyzed \citep{kveton2019garbage}. Generally, in our context, we generate multiple models ${ \hat r_1,\cdots,\hat r_M}$ by resampling datasets, and then consider $\argmin_{i} \hat r_i$ as $\hat r-\hat g$.
\end{example}

\section{Conservative Fine-tuning of Diffusion Models}\label{sec:conservative}

\begin{algorithm}[!t]
\caption{\textbf{\alg}\,(dou\textbf{B}ly conse\textbf{R}v\textbf{A}tive f\textbf{I}ne-tuning \textbf{D}iffusion models) }\label{alg:main_braid}
\begin{algorithmic}[1]
  \STATE {\bf Require}: Parameter $\alpha \in \RR^{+}$, a set of policy classes $\{\Pi_t\}$ where $\Pi_t \subset [\Xcal \to \Delta(\Xcal)]$, pre-trained diffusion model $\{p^{\pre}_t\}_{t=T+1}^1$. 
      \STATE Train a conservative reward model $\hat r-\hat g$ using an offline data $\Dcal_{\mathrm{off}}$.  \label{lst:step2}   
      \STATE  \label{lst:step3} Update a diffusion model as $\{\hat p_t\}_t$ by solving the planning problem: 
      \begin{align}\label{eq:key_plnanning}
       \{\hat p_t\}_t=   & \argmax_{\{p_t \in \Pi_t\}_{t=T+1}^1 }\underbrace{\EE_{\{p_t\}}[\hat r(x_0) - \hat g(x_0)]}_{\text{Penalized reward}}   - \alpha \underbrace{\Sigma_{t=T+1}^1  \EE_{\{p_t\}}[\mathrm{KL}(p_t(\cdot|x_t)\|p^{\pre}_t (\cdot |x_t))  ]}_{\text{KL\,penalty} }
      \end{align}
      where the expectation $\EE_{\{p_t\} }[\cdot]$ is taken with respect to $\prod_{t=T+1}^1 p_t(x_{t-1}|x_{t})$. 
  \STATE {\bf Output}: A policy $\{\hat p_t\}_t$   
\end{algorithmic}
\end{algorithm}

In this section, we consider how to sample from a doubly conservative generative model $\hat \pi_{\alpha}$, using diffusion models as pre-trained generative models. Our algorithm is outlined in Algorithm~\ref{alg:main_braid}. Initially, we learn a penalized reward $\hat r-\hat g$ from the offline data and set it as a target to prevent overoptimization in \eqref{eq:key_plnanning}. Additionally, we integrate a KL regularization term to prevent invalid designs. The parameter $\alpha$ governs the intensity of this regularization term.

Formally, this phase can be conceptualized as a planning problem in soft-entropy-regularized MDPs \citep{neu2017unified,geist2019theory}. In this MDP formulation:
\begin{itemize}
    \item The state space $\Scal$ and action space $\Acal$ correspond to the design space $\Xcal$.
    \item The reward at time $t \in[0,\cdots,T]$ ($\in\Scal \times \Acal \to \RR$) is provided only at $T$ as $\hat r- \hat g$.
    \item The transition dynamics at time $t$ ($\in[\Scal \times \Acal \to \Delta(\Scal)]$) is an identity $\delta(s_{t+1}={a_t})$.
    \item The policy at time $t$ ($\in \Scal \to \Delta(\Acal)$) corresponds to $p_{T+1-t}:\Xcal \to \Delta(\Xcal)$.
    \item The reference policy at $t$  is a policy in the pre-trained model $p^{\pre}_{T+1-t}$ 
\end{itemize}
In these entropy-regularized MDPs, the soft optimal policy corresponds to $\{\hat p_t\}$. Importantly, we can analytically derive the fine-tuned distribution in \pref{alg:main_braid} and show that it simplifies to a doubly conservative generative model $\hat \pi_{\alpha}$, from which we aim to sample.

\begin{theorem}\label{thm:key}
Let $\hat p_{\alpha}(\cdot)$ be an induced distribution from optimal policies $\{\hat p_t\}_{t= T+1}^1$ in \eqref{eq:key_plnanning}, i.e., $\hat p(x_0) = \int \{\prod_{t=T+1}^1 \hat p_t(x_{t-1}|x_t)\}d x_{1:T}$ when $\{\Pi_t\}$ is a global policy class ($\Pi_t= \{\Xcal \to \Delta(\Xcal)\}$). Then,   
\begin{align*}
    \hat p_{\alpha}(x)= \hat \pi_{\alpha}(x).  
\end{align*}
\end{theorem}  

We have deferred to the proof in Section~\ref{subsec:proof}. While similar results are known in the context of standard entropy regularized RL \citep{levine2018reinforcement}, our theorem is novel because previous studies did not consider pre-trained diffusion models.

\paragraph{Training algorithms.}

Based on Theorem~\ref{thm:key}, to sample from $\hat \pi_{\alpha}$, what we need to is to solve Equation~\eqref{eq:key_plnanning}. We can employ any off-the-shelf RL algorithms to solve this planning problem. Given that the transition dynamics are known, and differentiable reward models are constructed in our scenario, a straightforward approach to optimize \eqref{eq:key_plnanning} is to directly optimize differentiable loss functions with respect to parameters of neural networks in policies, as detailed in Appendix~\ref{sec:direct}. Indeed, this approach has recently been used in fine-tuning diffusion models \citep{clark2023directly,prabhudesai2023aligning}, demonstrating its stability and computational efficiency.

\begin{remark}[Novelty of \pref{thm:key}]\label{rem:difference}
A theorem similar to \pref{thm:key} has been proven for continuous-time diffusion models in Euclidean space \citep[Theorem 1]{uehara2024fine}. However, the primary distinction lies in the fact that while their findings are restricted to Euclidean space, where diffusion policies take Gaussian polices, our results are not constrained to any specific domain. Hence, for example, our \pref{thm:key} can handle scenarios where the domain is discrete or lies on the simplex space \citep{avdeyev2023dirichlet} in order to model biological sequences as we do in \pref{sec:main_experiments}. 
\end{remark}

\subsection{Sketch of the Proof of Theorem~\ref{thm:key}} 
\label{subsec:proof} 

We explain the sketch of the proof of Theorem~\ref{thm:key}. The detail is deferred to \pref{thm:key_proof}. 

By induction from $t=0$ to $t=T+1$, we can first show  
\begin{align}\label{eq:optimal_i}\textstyle 
     \hat p_t(x_{t-1}|x_t)= \frac{\exp(v_{t-1}(x_{t-1})/\alpha)p^{\pre}_{t-1}(x_{t-1}|x_t)}{\exp( v_{t}(x_t)/\alpha)  }. 
\end{align}
Here, $v_{t}(x_{t})$ is a soft optimal value function: 
   \begin{align*}\textstyle 
      \EE_{\hat p}[\hat r(x_0) - \hat g(x_0) - \alpha  \sum_{k=t}^1 \mathrm{KL}(p_k(\cdot|x_k)\|p^{\pre}_k (\cdot |x_k))  |x_t ], 
 \end{align*}
which satisfies an equation analogous to the soft Bellman equation: $ v_{0}(x) = \hat r(x) - \hat g(x)$ and for $t=1$ to $t=T+1$,
\begin{align}\textstyle \label{eq:soft} 
 \exp(\frac{ v_{t}(x_{t})}{\alpha})= \int \exp\left (\frac{ v_{t-1}(x_{t-1})}{\alpha}\right)p^{\pre}_{t}(x_{t-1} \mid x_{t}) d x_{t-1}. 
\end{align}

Now, we aim to calculate a marginal distribution at $t$ defined by $\hat p_t(x_t)=\int \{\prod^t_{k=t+1} \hat p_{k}(x_{k-1}|x_{k})\}d x_{t+1:T}$. Then, by induction, we can show that 
\begin{align}\label{eq:marginal}\textstyle 
    \hat p_t(x_t)= \exp(v_{t}(x_{t})/\alpha)  p^{\pre}_t(x_t)/C 
\end{align}
where $C$ is a normalizing constant. Indeed, supposing that the above \eqref{eq:marginal} hold at $t$, the equation \eqref{eq:marginal} also holds for $t-1$ as follows: 
\begin{align*}\textstyle 
    \int  \hat p_{t-1}(x_{t-1} |x_{t}) \hat p_t(x_t )d x_t = \exp(v_{t-1}(x_{t-1})/\alpha)  p^{\pre}_{t-1}(x_{t-1})/C = \hat p_{t-1}(x_{t-1}). 
\end{align*}
Finally, by setting $t=0$, the statement in \pref{thm:key} is concluded.

\section{Regret Guarantee}
In this section, our objective is to demonstrate that a policy $\hat p_{\alpha}$ from \pref{alg:main_braid} can provably outperform designs in offline data by establishing the regret guarantee. 

To assess the performance of our fine-tuned generative model, we introduce the soft-value metric:
\begin{align*}
    J_{\alpha}(p):= \EE_{x \sim p}[r(x)] - \alpha \KL(p\|  p_{\pre}).  
\end{align*}
This metric comprises two components: the expected reward and a penalty term applied when $p$ produces invalid outputs, as we see in \pref{eq:goal_objective}. Now, in terms of soft-value $J_{\alpha}(p)$, our proposal $\hat p_{\alpha} $ offers the following guarantee.

\begin{theorem}[Per-step regret]\label{thm:regret}
Suppose Assumption~\ref{assum:crucial}. Then, with probability $1-\delta$, we have 
\begin{align*} \textstyle 
   \forall \pi \in \Delta(\Xcal); \underbrace{J_{\alpha}(\pi) - J_{\alpha}(\hat p_{\alpha})}_{\textit{Per step regret }}  & \leq 2 \sqrt{C_{\pi}}\times  \underbrace{\EE_{x \sim p_{\mathrm{off}}}[\hat g(x)^2]^{1/2} }_{\textit{Stat}}, \quad  C_{\pi}:= \max_{x \in \Xcal_{\mathrm{off}}}  \left | \frac{\pi(x)}{p_{\mathrm{off}}(x)} \right |, 
\end{align*}
where $\Xcal_{\mathrm{off}}=\{x \in \Xcal: p_{\mathrm{off}}(x)>0\}$. As an immediate corollary, 
\begin{align*} \textstyle 
   \EE_{x\sim \pi}[r(x)] - \EE_{x \sim \hat p_{\alpha} }[r(x)]  & \leq \alpha \KL(\pi\|p_{\pre})  + 2  \sqrt{C_{\pi}}\times  \EE_{x \sim p_{\mathrm{off}}}[\hat g(x)^2]^{1/2}. 
\end{align*}
\end{theorem}

In the theorem above, we establish that the per-step regret against a generative model $\pi$ we aim to compete with is small as long as the generative model $\pi$ falls within $\Xcal_{\mathrm{off}}$ and the learned model $\hat r$ is calibrated as in Assumption~\ref{assum:crucial}. First, the term (Stat) corresponds to the statistical error associated with $\hat r$ over the offline data distribution $p_{\mathrm{off}}$. When the model is well-specified, it is upper-bounded by $\sqrt{\bar d/n}$, where $\bar d$ represents the effective dimension of $\Fcal$, as we will discuss shortly. Secondly, the term $C_{\pi}$ corresponds to the coverage between a comparator generative model $\pi$ and our generative model $\hat p_{\alpha} $. Hence, it indicates that the performance of our learned $\hat p_{\alpha} $ is at least as good as that of a comparator generative model $\pi$ covered by $p_{\mathrm{off}}$. While this original coverage term $C_{\pi}$ diverges when $\pi$ goes outside of $\Xcal_{\mathrm{off}}$, we can refine it using the extrapolation capabilities of a function class $\Fcal$, as we will discuss shortly. This refined version ensures that we can achieve high-quality designs that outperform designs in the offline data (i.e., best designs in $\Xcal_{\mathrm{off}}$).

\begin{example}

We consider a scenario where an RKHS is used for $\Fcal$. Let $\Fcal$ be a model represented by an infinite-dimensional feature $\phi(\cdot)$. Let $\bar d$ denote the effective dimension of $\Fcal$ \citep{valko2013finite}.

\begin{corollary}[Informal: Formal characterization is in \pref{sec:GPs} ]\label{cor:GPS}
Assuming that the model is well-specified, with probability $1-\delta$, we have:
\begin{align*}\textstyle 
    J_{\alpha}(\pi) - J_{\alpha}(\hat p_{\alpha} )\leq  \sqrt{\bar C_{\pi}} \times \tilde O\left(\sqrt{ \frac{\bar d^{3}}{n}} \right),\quad \bar C_{\pi}:=\sup_{\kappa:\|\kappa \|_2=1 }\frac{\kappa ^{\top}\EE_{x\sim \pi}[\phi(x) \phi^{\top}(x) ] \kappa}{\kappa ^{\top} \EE_{x\sim p_{\mathrm{off}}}[\phi(x) \phi^{\top}(x)] \kappa}. 
\end{align*}

The refinement of the coverage term in $\bar C_{\pi}$ is characterized as the relative condition number between covariance matrices on a generative model $\pi$ and an offline data distribution $p_{\mathrm{off}}$, which is smaller than $C_{\pi}$. This $\bar C_{\pi}$ could still be finite even if $C_{\pi}$ is infinite. In this regard, Corollary~\ref{cor:GPS} illustrates that the trained generative model can outperform the best design in the offline data by harnessing the extrapolation capabilities of reward models.
\end{corollary}

\end{example}

\section{Experiments} \label{sec:main_experiments}

We perform experiments to evaluate (a) the effectiveness of conservative methods for fine-tuning diffusion models and (b) the comparison of our approach between existing methods for MBO with diffusion models \citep{krishnamoorthy2023diffusion,yuan2023reward}. We will start by outlining the baselines and explaining the experimental setups. Regarding more detailed setups, hyperparameters, architecture of neural networks, and ablation studies, refer to Appendix~\ref{sec:experiments}. 

\paragraph{Methods to compare.}  We compare the following methods in our evaluation. For a fair comparison, we always use the same $\alpha$ in \textbf{\alg} and \textbf{STRL}. 
\footnote{Regarding the effectiveness of KL-regularization, it has been discussed in \citet{fan2023dpok,uehara2024fine}. Hence, in our work, we focus on the effectiveness of conservatism in reward modeling. }. 

\begin{itemize}
    \item \textbf{\alg~(proposed method)}: We consider two approaches: (1) \textbf{Bonus}, as in Example~\ref{exa:gps} by setting a last layer as a feature map and constructing a kernel, (2) \textbf{Bootstrap}, as in Example~\ref{exa:bootstrap}. 
    \item \textbf{Standard RL (STRL)}: RL-fine-tuning that optimizes the standard $\hat r$ without any conservative term, following existing works on fine-tuning \citep{clark2023directly,prabhudesai2023aligning}. 
    \item \textbf{DDOM \citep{krishnamoorthy2023diffusion}}: We train with weighted classifier-free guidance \citep{ho2022classifier} using offline data, conditioning on a class with high $y$ values (top $5\%$) during inference. Note that this method is training from scratch rather than fine-tuning.
    \item \textbf{Offline Guidance \citep{yuan2023reward}}: After training a classifier using offline data, we use guidance (conditional diffusion models) \citep{dhariwal2021diffusion} on top of pre-trained diffusion models and condition on classes with high $y$ values (top $5\%$) at inference time.
\end{itemize} 

\paragraph{Evaluation.} We assess the performance of each generative model primarily by visualizing the histogram of true rewards $r(x)$ obtained from the generated samples. For completeness, we include similar histograms for both the pre-trained model (\textbf{Pretrained}) and the offline dataset (\textbf{Offline}). As for hyperparameter selection, such as determining the strengths of conservative terms/epochs, we adhere to conventional practice in offline RL (e.g., \citet{rigter2022rambo,kidambi2020morel,matsushima2020deployment}) and choose the best one through a limited number of online interactions.

\begin{remark} We omit comparisons with pure MBO methods for two reasons: (i) \textbf{DDOM}, which we compare against, already demonstrates a good performance across multiple datasets, and (ii) these methods are unable to model complex valid spaces since they do not incorporate state-of-the-art generative models (e.g., stable diffusion), thereby lacking the capability to generate valid designs (e.g., natural images) as we show in \pref{sec:images}. 
\end{remark}

\subsection{Design of Regulatory DNA/RNA Sequences}\label{subsec:regulatory}

\begin{figure}[!t]
    \centering    
    \begin{subfigure}{.32\textwidth}
    \includegraphics[width = \textwidth]{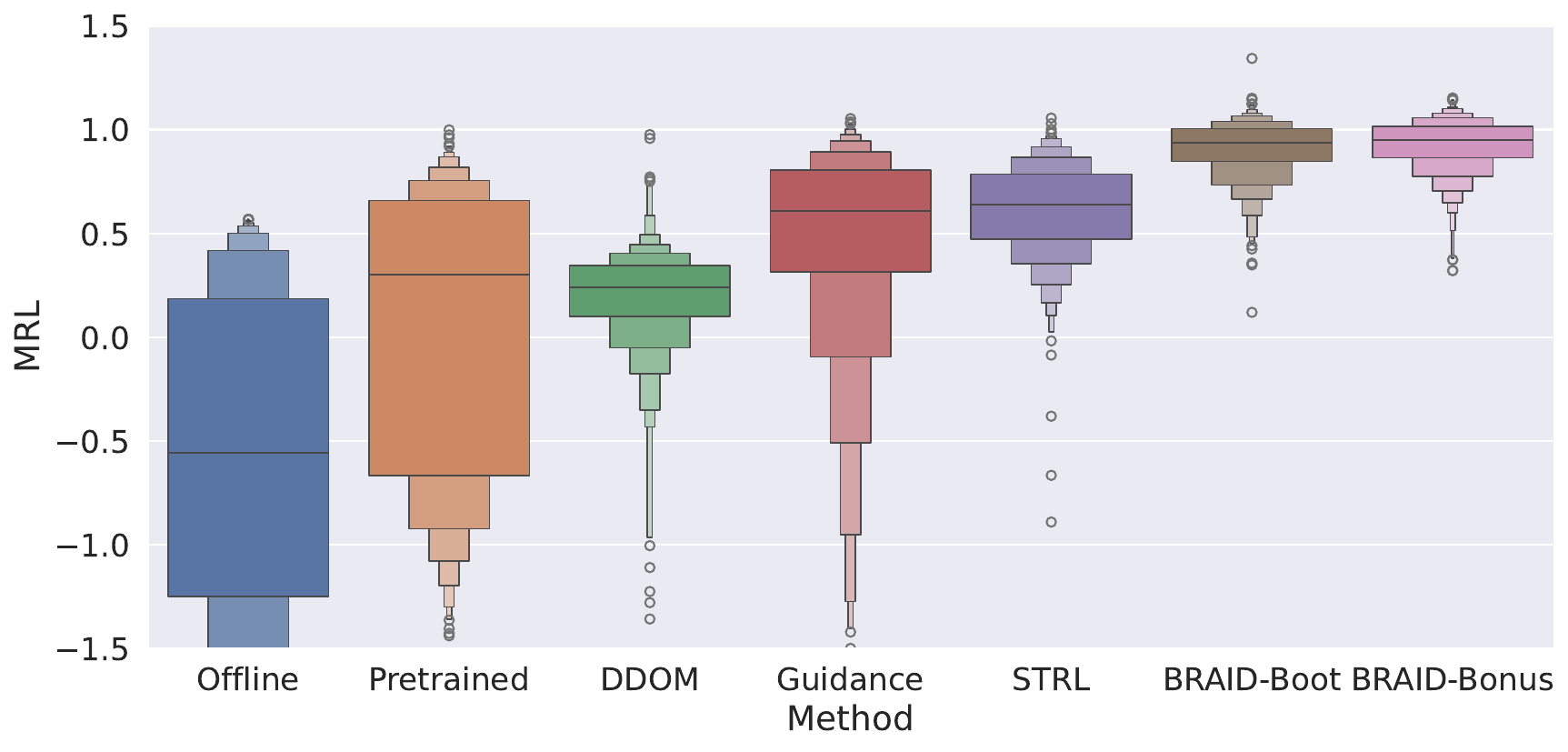}
    \caption{5'UTRs}
    \end{subfigure}
    \begin{subfigure}{.32\textwidth}
    \includegraphics[width = \textwidth]{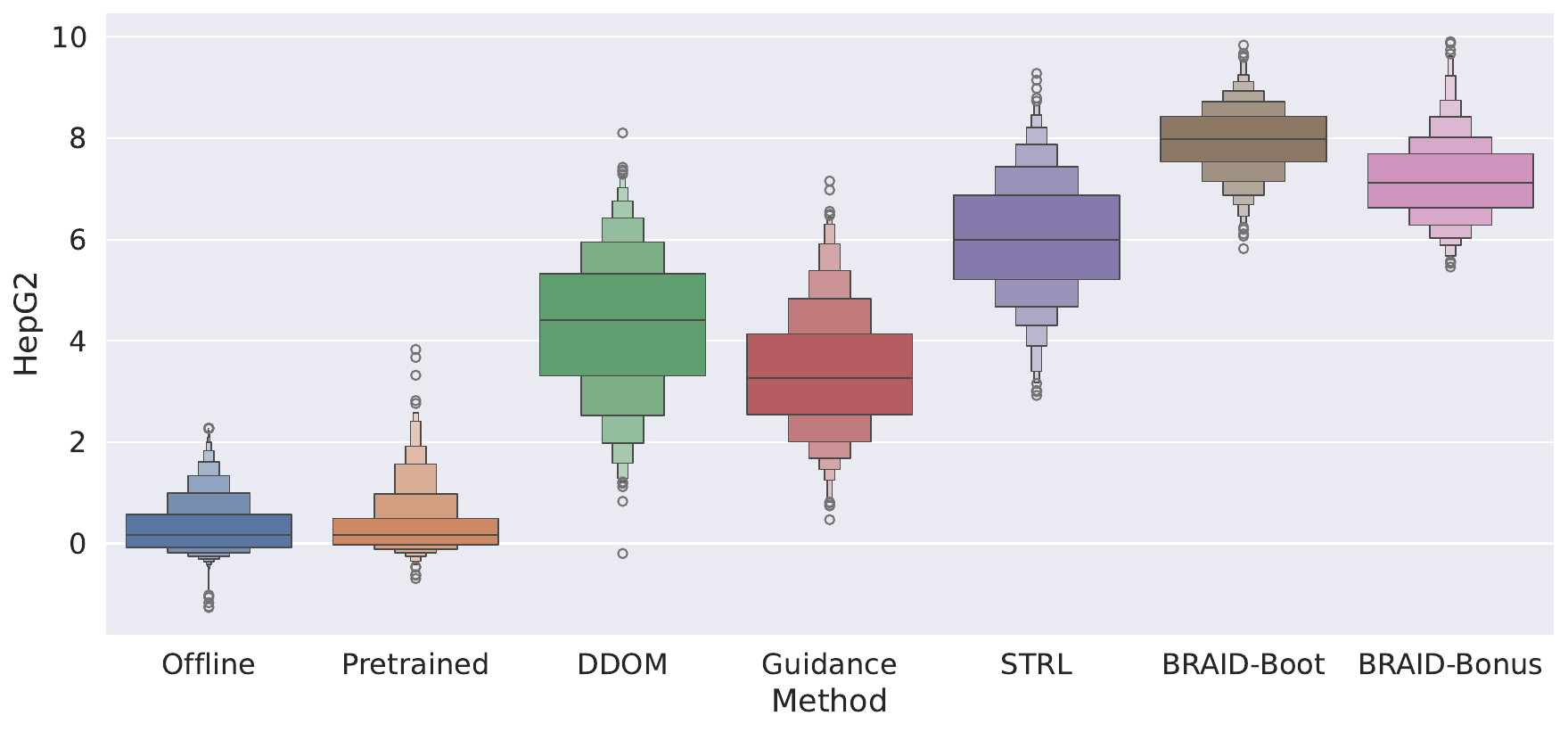}
    \caption{Enhancers}
    \end{subfigure} 
    \begin{subfigure}{.32\textwidth}
    \includegraphics[width = \textwidth]{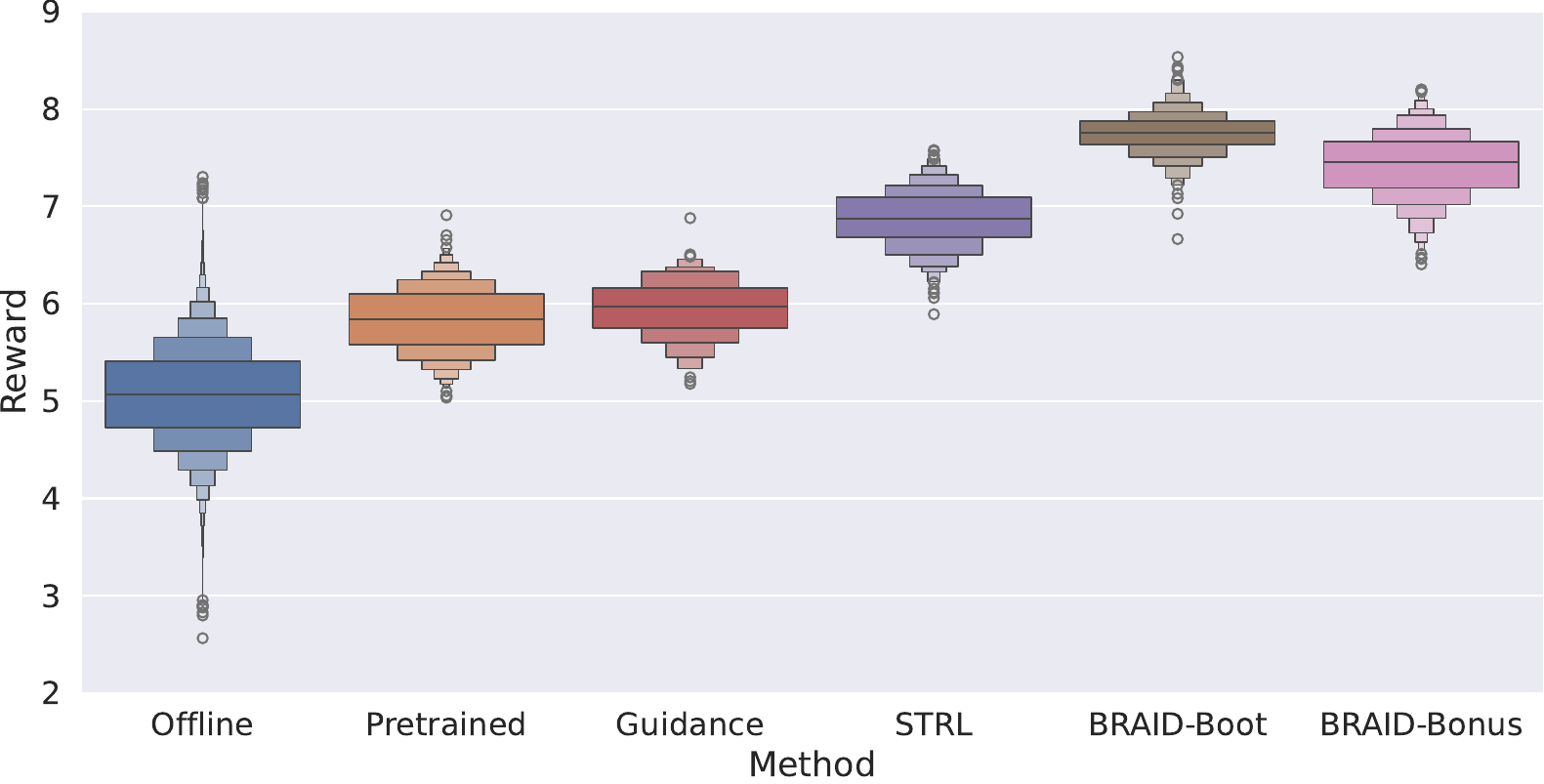}
    \caption{Images}
    \end{subfigure} 
    \caption{Barplots of the rewards \emph{$r(x)$} for samples generated by each algorithm. It reveals that proposals consistently outperform baselines. 
    }  
    \label{fig:bio_images}
\end{figure}

We examine two publicly available large datasets consisting of enhancers ($n \approx 700k $) \citep{gosai2023machine} and UTRs ($n \approx 300k$) \citep{sample2019human} with activity levels collected by massively parallel reporter assays (MPRA) \citep{inoue2019identification}. These datasets have been extensively used in sequence optimization for DNA and RNA engineering, particularly for the advancement of cell and RNA therapy \citep{castillo2021machine,ghari2023generative,lal2024reglm,ferreira2024dna}. In the Enhancers dataset, each $x$ is a DNA sequence with a length of $200$, while $y \in \RR$ is the measured activity in cell lines. For the UTRs dataset, $x$ is a 5'UTR RNA sequence with a length of $50$ while $y \in \RR$ is the mean ribosomal load (MRL) measured by polysome profiling.

\paragraph{Setting of oracles and offline data.}  We aim to explore a scenario where we have a pre-trained model and an offline dataset. Since the true reward function $r(\cdot)$ is typically unknown, we initially divide the original dataset $\mathcal{D}=\{x^{(i)},y^{(i)}\}$ randomly into two subsets: $\mathcal{D}_{\text{ora}}$ and $\mathcal{D'}$. Then, from $\mathcal{\Dcal'}$, we select datasets below $95\%$ quantiles for enhancers and $60\%$ quantiles for UTRs and define them as offline datasets $\mathcal{D}_{\text{off}}$. Subsequently, we construct an oracle $r(\cdot)$ by training a neural network on $\mathcal{D}_{\text{ora}}$ and use it for testing purposes. Here, we use an Enformer-based model, which is a state-of-the-art model for DNA sequences \citep{avsec2021effective}. Regarding pre-trained diffusion models, we use ones customized for sequences over simplex space \citep{avdeyev2023dirichlet}. 
In the subsequent analysis, each algorithm solely has access to the offline data $\mathcal{D}_{\text{off}}$ and a pre-trained diffusion model, but not $r(\cdot)$.

\paragraph{Results.} The performance results in terms of $r(\cdot)$ are depicted in Fig~\ref{fig:bio_images}a and b. It is seen that fine-tuned generative models via RL outperform conditioning-based methods: \textbf{DDOM} and \textbf{Guidance}. This is expected because conditional models themselves are not originally intended to surpass the conditioned value ($\approx$ best value in the offline data). Conversely, fine-tuned generative models via RL are capable of exceeding the best value in offline data by harnessing the extrapolation capabilities of reward modeling, as also theoretically supported in Corollary~\ref{cor:GPS}. Secondly, both \textbf{BRAID-boot} and \textbf{BRAID-bonus} demonstrate superior performance compared to \textbf{STRL}.%
This suggests that conservatism aids in achieving fine-tuned generative models with enhanced rewards while mitigating overoptimization.

\subsection{Image Generation}\label{sec:images}

\begin{wrapfigure}{r}{0.42\textwidth}
\centering 
\begin{subfigure}{.16\textwidth}
  {\includegraphics[width=\textwidth]{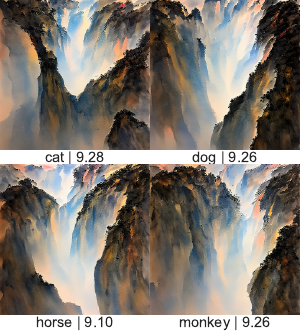}
  \caption{Undesitable scenarios }  \label{fig:undesirable}}
\end{subfigure}
\begin{subfigure}{.21\textwidth}
  {\includegraphics[width=\textwidth]{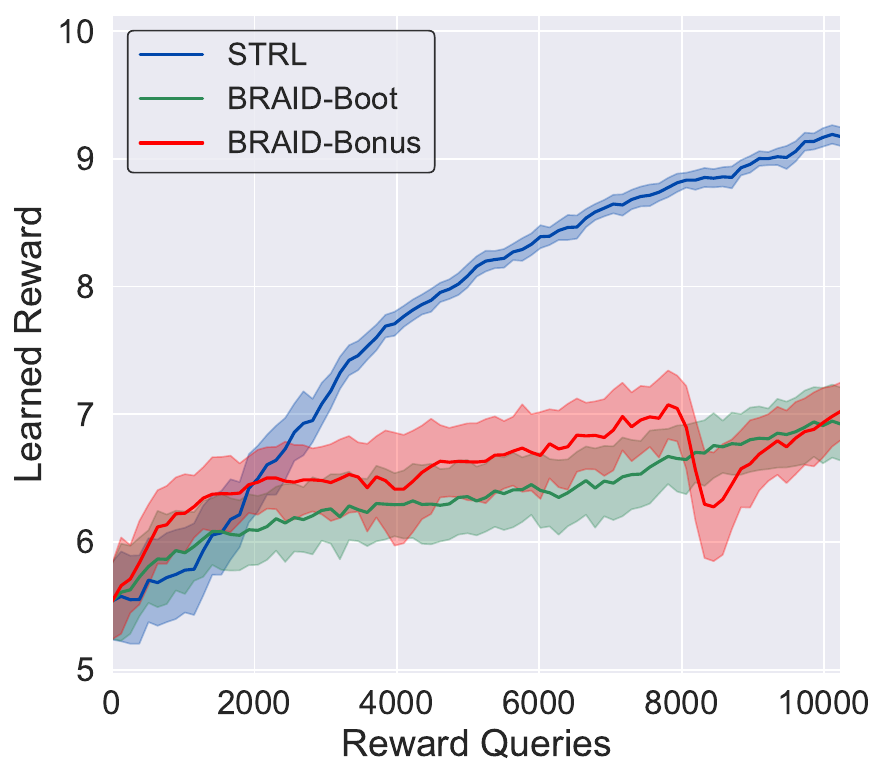}
  \caption{Training curves in terms of $\hat r$ (but not $r$) }  \label{fig:training}}
\end{subfigure}
    {\includegraphics[width=0.40\textwidth]
    {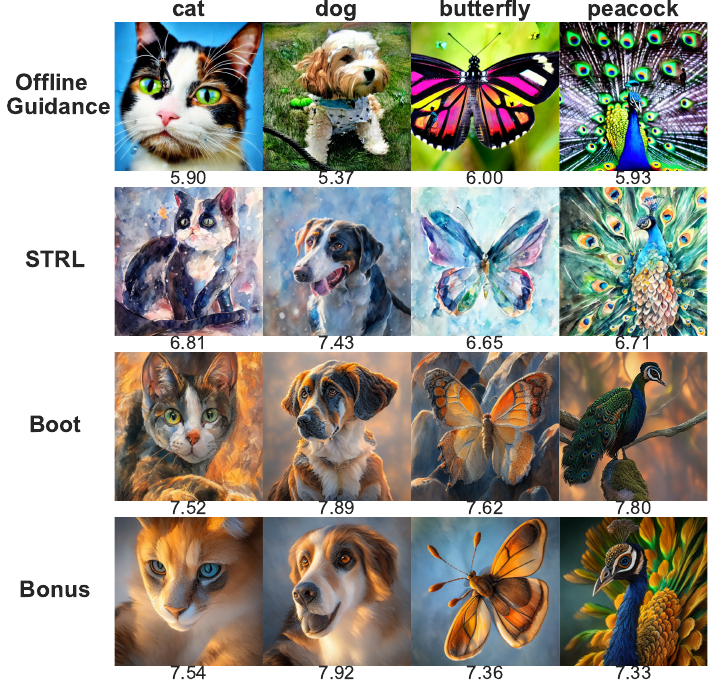}
  \caption{(c) Generated images }
   \label{fig:generated_images}}
\end{wrapfigure}

We consider the task of generating aesthetically pleasing images, following prior works \citep{fan2023dpok,black2023training}. We use Stable Diffusion v1.5 as our pretrained diffusion model, which can generate high-quality images conditioned on prompts such as ``cat'' and  ``dog''. We use the AVA dataset \citep{murray2012ava} as our offline data and employ a linear MLP on top of CLIP embeddings to train reward models ($\hat r$ and $\hat r-\hat g$) from offline data for fine-tuning.

\paragraph{Setting of oracles.}
To construct $r(x)$, following existing works, we use the LAION Aesthetic Predictor V2 \citep{schuhmann2022laion}, already pre-trained on a large-scale image dataset. However, this LAION predictor gives high scores even if generated images are almost identical regardless of prompts, as in Figure~\ref{fig:undesirable}. These situations are undesirable because it means fine-tuned models are too far away from pre-trained models. Hence, for our evaluation, we define $r(x)$ as follows: (1) asking vision language models (e.g., LLaVA~\citep{liu2024visual}) whether images contain objects in the original prompts \footnote{The F1 score of LLaVA for detecting objects was $1.0$ as detailed in Appendix~\ref{subsec:LLAVA_guided}. Hence, this part is considered to be accurate.} (e.g., dog, cat), (2) if Yes, outputting the LAION predictor, and (3) if No, assigning $0$. This evaluation ensures that high $r(x)$ still indicates capturing the space of the original stable diffusion.

\paragraph{Results.} We show that our proposed approach outperforms the baselines \emph{in terms of $r(x)$}, as in Fig~\ref{fig:bio_images}c \footnote{Note we omitted \textbf{DDOM} because it is not a fine-tuning method.}. We also show the generated images in Figure~\ref{fig:generated_images}c. Additionally, we plot the training curve during the fine-tuning process \emph{in terms of the mean of $\hat{r}(x)$} of generated samples in Fig~\ref{fig:generated_images}b. The results indicate that in \textbf{STRL}, while the learning curve based on the learned reward quickly grows, fine-tuned models no longer necessarily remain within the space of pre-trained models (\textbf{STRL} in Fig \ref{fig:bio_images}c). In contrast, in our proposal, by carefully regularizing on regions outside of the offline data, we can generate more aesthetically pleasing images than \textbf{STRL}, which remain within the space of pre-trained models. For more images/ ablation studies, refer to Appendix~\ref{sec:experiments}.

\vspace{-2mm}
\section{Summary}\label{sec:summary}
\vspace{-2mm}

For the purpose of fine-tuning from offline data, we introduced a conservative fine-tuning approach by optimizing a conservative reward model, which includes additional penalization outside of offline data distributions. Through empirical and theoretical analysis, we demonstrate the capability of our approach to outperform the best designs in offline data, leveraging the extrapolation capabilities of reward models while avoiding the generation of invalid designs through pre-trained diffusion models.

\bibliographystyle{chicago}
\bibliography{rl}

\appendix 

\newpage 

\onecolumn

\section{Additional Related Works}\label{sec:additional}

In this section, we summarize additional related works. 

\paragraph{Conservative approaches in offline RL/offline contextual bandits. }

Conservative approaches have been explored in offline RL and contextual bandits. Firstly, in both model-free and model-based RL, one prevalent method involves incorporating an additional penalty on top of the reward functions \citep{yu2020mopo,chang2021mitigating}. Secondly, in model-based RL, a common strategy is to train transition dynamics in a conservative manner \citep{kidambi2020morel,rigter2022rambo,uehara2021pessimistic}. Thirdly, in model-free RL, typical approaches include conservative learning of q-functions \citep{kumar2020conservative,xie2021bellman} or the inclusion of KL penalties against behavioral policies \citep{wu2019behavior,fakoor2021continuous}.

However, these works are not designed to incorporate a diffusion model, unlike our approach. Hence, it remains unclear how their works can generate designs that remain within intricate valid design space, such as high-quality images using stable diffusion.

\paragraph{Design with generative models.}

Many works are focusing on design problems with generative models. However, these works are typically limited to the usage of VAEs. \citep{notin2021improving,gomez2018automatic} Our work is still significantly different because we focus on generative models.

\paragraph{Fine-tuning in LLMs from human feedbacks.}

A closely related area of research involves fine-tuning LLMs through the optimization of reward functions using human feedback \citep{touvron2023llama,ouyang2022training}. Especially, following works such as \citet{zhan2023provable}, from a theoretical viewpoint, \citet{xiong2023iterative} explores the effectiveness of pessimism in offline scenarios and its theoretical aspect. However, our theoretical findings are more specifically tailored to diffusion models. Indeed, our main result in Theorem~\ref{thm:key} is novel, and our algorithm differs significantly, as fine-tuning methods in offline settings in the literature on LLMs typically rely on policy gradient or PPO, whereas we use more direct backpropagation approaches. Furthermore, the meaning of step size is different as well. Hence, in \citep{xiong2023iterative}, they do not use soft entropy regularized MDPs. 

Typically, in the above works, human feedback is considered to be given in the form of preferences. Similarly, in the context of diffusion models, \citet{yang2023using,wallace2023diffusion} discusses fine-tuning of diffusion models using preference-based feedback. However, these works focus on online settings but not offline settings.  
 
\paragraph{Sampling from unnormalized distributions.}

In our approach, we use an RL method to sample from our target distribution that is proportional to $\exp((\hat r-\hat g(x))p_{\pre}(x)$.  While MCMC has traditionally been prevalent in sampling from unnormalized Boltzmann distributions $\exp(r(x))$, recent discussions \citep{zhang2021path,vargas2023denoising}, have explored RL approaches similar to our approach. However, their focus differs from ours, as they do not address the fine-tuning of diffusion models (i.e., no $p_{\pre}$) or sample efficiency in offline settings.

Another relevant literature discusses sampling from unnormalized Boltzmann distributions when pre-trained diffusion models are available \citep{kong2024diffusion}. However, their algorithm closely resembles classifier-based guidance \citep{dhariwal2021diffusion}, rather than a fine-tuning algorithm. Additionally, they do not examine conservatism in an offline setting. 

\section{Direct Back Propagation}\label{sec:direct}

Our planning algorithm has been summarized in \pref{alg:main}. Here, we parametrize each policy by neural networks. 

For the sake of explanation, we also add typical cases where the domain is Euclidean in \pref{alg:main_eculidian}.

\begin{algorithm}[!h]
\caption{Direct Back Propagation (General case)  }\label{alg:main}
\begin{algorithmic}[1]
     \STATE {\bf Require}: Set a diffusion-model $p(\cdot|x_{t-1};\theta)$, pre-trained model $\{p_{\pre}(\cdot|x_{t-1}) \}_{t=T+1}^1$, batch size $m$, a parameter $\alpha \in \RR^+$. 
     \STATE {\bf Initialize}: $\theta_1 = \theta_{\pre}$
     \FOR{$s \in [1,\cdots,S]$}
      \STATE Set $\theta = \theta_s$. 
      \STATE Collect $m$ samples $\{x^{(i)}_t(\theta) \}_{t=T+1}^0$ from a current diffusion model (i.e., generating by sequentially running polices $\{p_t(\cdot|x_{t};\theta) \}_{t=T+1}^1$ from $t=T+1$ to $t=1$) 
      \STATE Update $\theta_s$ to $\theta_{s+1}$ by adding the gradient of the following loss $L(\theta)$ with respect to $\theta$ at $\theta_s$: 
      \begin{align}\label{eq:key}
   & L(\theta) = \frac{1}{m} \sum_{i=1}^m  \left[\hat r(x^{(i)}_0(\theta) ) - \hat g(x^{(i)}_0(\theta) ) -  \alpha     \sum_{t=T+1}^1 \KL(p_t(\cdot|x_{t};\theta)   \| p_{\pre}(\cdot|x_{t})  )  \right ] . 
      \end{align}  
      \ENDFOR
  \STATE {\bf Output}: Policy $\{p_t(\cdot|\cdot; \theta_S) \}_{t=T+1}^1$   
\end{algorithmic}
\end{algorithm}

\begin{algorithm}[!h]
\caption{Direct Back Propagation (in Euclidean space) }\label{alg:main_eculidian}
\begin{algorithmic}[1]
     \STATE {\bf Require}: Set a diffusion-model $\{\Ncal(\rho(t,x_{t};\theta),\sigma^2_t); \theta \in \Theta\}_{t=T+1}^1$, pre-trained model $\{\Ncal(\rho(t,x_{t};\theta_{\pre}),\sigma^2_t)\}_{t=T+1}^1$, batch size $m$, a parameter $\alpha \in \RR^+$. 
     \STATE {\bf Initialize}: $\theta_1 = \theta_{\pre}$
     \FOR{$s \in [1,\cdots,S]$}
      \STATE Set $\theta = \theta_s$. 
      \STATE Collect $m$ samples $\{x^{(i)}_t(\theta) \}_{t=T+1}^0$ from a current diffusion model (i.e., generating by sequentially running polices $\{\Ncal(\rho(t,x_{t};\theta),\sigma^2_t)\}_{t=T+1}^1$ from $t=T+1$ to $t=1$) 
      \STATE Update $\theta_s$ to $\theta_{s+1}$ by adding the gradient of the following loss $L(\theta)$ with respect to $\theta$ at $\theta_s$: 
      \begin{align}\label{eq:key}
   & L(\theta) = \frac{1}{m} \sum_{i=1}^m  \left[\hat r(x^{(i)}_0(\theta) ) - \hat g(x^{(i)}_0(\theta) ) -  \alpha     \sum_{t=T+1}^1 \frac{\|\rho(x^{(i)}_t(\theta),t;\theta)-\rho(x^{(i)}_t(\theta),t; \theta_{\pre}) \|^2 }{2\sigma^2(t) } \right ] . 
      \end{align}  
      \ENDFOR
  \STATE {\bf Output}: Policy $\{p_t(\cdot|\cdot; \theta_S) \}_{t=T+1}^1$   
\end{algorithmic}
\end{algorithm}

\section{All Proofs}

\subsection{Proof of Theorem~\ref{thm:key}}\label{thm:key_proof}

Here, we actually prove a stronger statement. 

\begin{theorem}[Marginal and Posterior distributions]\label{thm:key2}
Let $\hat p_t(x_t)$ and $\hat p^b_{t}(x_{t}| x_{t-1})$ be  marginal distributions at $t$ or posterior distributions of $x_{t}$ given $x_{t-1}$, respectively, induced by optimal policies $\{\hat p_t\}_{T+1}^1$. Then, 
\begin{align*}
    \hat p_t(x_t) =  \exp(v_{t}(x_{t})/\alpha)\hat p^{\pre}_{t}(x_{t})/C,\quad \hat p^b_t(x_{t}|x_{t-1}) = \hat p^{\pre}_t(x_{t}|x_{t-1}). 
\end{align*}
\end{theorem}

\paragraph{Proof.}  To simplify the notation, we let $f(x) =  \hat r(x)- \hat g(x)$. 
As a first step, by using induction, we aim to obtain an analytical form of the optimal policy $\{\hat p_t(\cdot|x_{t-1})\}$. 

First, we define the soft-optimal value function as follows: 
\begin{align*}
v_{t-1}(x_{t-1})  = \EE_{\{\hat p_t\} }\left[f(x)  - \alpha \sum_{k=t-1}^1 \KL(\hat p_k(\cdot|x_k) \|p^{\pre}_k(\cdot|x_k)) | x_{t-1} \right]. 
\end{align*}
Then, by induction, we have 
\begin{align*}
 \hat p_t(x_{t-1}|x_t) =\argmax_{ p_t \in \Delta(\Xcal) } \EE_{\{\hat p_t\} }\left [ v_{t-1}(x_{t-1}) - \alpha \KL(p_t(\cdot|x_t) \|p^{\pre}_t(\cdot|x_t)) |x_t \right].   
\end{align*}
With simple algebra, we obtain 
\begin{align}\label{eq:proportional}
\hat p_t(x_{t-1}|x_t) \propto \exp\left(\frac{v_{t-1}(x_{t-1}) }{\alpha} \right)  p^{\pre}_t(x_{t-1}|x_t) .  
\end{align}
Here, noting 
\begin{align*}
 v_t(x_t) =\max_{p_t \in \Delta(\Xcal) } \EE_{\{\hat p_t\} }[ v_{t-1}(x_{t-1}) - \alpha \KL(p_t(\cdot|x_t) \|p^{\pre}_t(\cdot|x_t)) |x_t ],
\end{align*}
we get the soft Bellman equation:
\begin{align}\label{eq:normalizing}
  \exp\left(\frac{v_{t}(x_{t}) }{\alpha} \right)  = \int \exp\left(\frac{v_{t-1}(x_{t-1}) }{ \alpha} \right)  p^{\pre}_t(x_{t-1}|x_t) \mathrm{d}x_{t-1}. 
\end{align}
Therefore, by plugging \eqref{eq:normalizing} into \eqref{eq:proportional},  we actually have 
\begin{align}\label{eq:soft_formulation}
 \hat p_t(x_{t-1}|x_t)=\frac{ \exp\left(\frac{v_{t-1}(x_{t-1}) }{\alpha} \right) p^{\pre}_t(x_{t-1}|x_t)}{  \exp\left(\frac{v_{t}(x_{t}) }{\alpha} \right)   } .
\end{align}

Finally, with the above preparation, we calculate the marginal distribution: 
\begin{align*}
    \hat p_t(x_t):=\int \left\{ \prod_{s=T+1}^t \hat p_{s}(x_{s-1}|x_s) \right\} dx_{t+1:T+1}. 
\end{align*}
Now, by using induction, we aim to prove 
\begin{align*}
    \hat p_{t}(x_{t})=  \exp\left(\frac{v_{t}(x_{t}) }{\alpha} \right)p^{\pre}_{t}(x_{t}). 
\end{align*}
Indeed, when $t=T+1$, from \eqref{eq:soft_formulation},  this hold as follows: 
\begin{align*}
    \hat p_{T+1}(x_{T+1}) = \frac{1}{C}  \exp\left(\frac{v_{T}(x_{T+1}) }{ \alpha} \right)p^{\pre}_{T+1}(x_{T+1}). 
\end{align*}
Now, suppose the above holds at $t$. Then, this also holds for $t-1$: 
\begin{align*}
      \hat p_{t-1}(x_{t-1})& =\int  \hat p_t(x_{t-1}|x_t)   \hat p_{t}(x_{t})\mathrm{d}x_t  \\
      &=\int \exp\left(\frac{v_{t-1}(x_{t-1}) }{\alpha} \right) \{ p^{\pre}_t(x_{t-1}|x_t) \} p^{\pre}_{t}(x_{t}) \mathrm{d}x_t  \tag{Use Equation~\ref{eq:soft_formulation}}  \\
      &= \exp\left(\frac{v_{t-1}(x_{t-1}) }{\alpha} \right)p^{\pre}_{t-1}(x_{t-1}). 
\end{align*}
By invoking the above when $t=0$, the statement is concluded.

\subsection{Proof of Theorem~\ref{thm:regret} }

In the following, we condition on the event where 
\begin{align*}
  \forall x\in \Xcal_{\pre};  |r(x) - \hat r(x)| \le g(x). 
\end{align*}
holds. 

First, we define 
\begin{align*}
    \hat J_{\alpha}(\pi) &:= \EE_{x \sim \pi}[ \hat r(x)-\hat g(x) ] -\alpha \mathrm{KL}(\pi \| p_{\un}),\quad J_{\alpha}(\pi):= \EE_{x \sim \pi}[ r(x)] -\alpha \mathrm{KL}(\pi \| p_{\un}). 
\end{align*}
We note that, $\hat{\pi}_\alpha$ maximizes $\hat{J}_\alpha(\pi)$. Therefore, we have 
\begin{align*}
       J_{\alpha}(\pi)- J_{\alpha}(\hat \pi_\alpha)&= J_{\alpha}(\pi)-  \hat J_{\alpha}(\pi)+ \hat J_{\alpha}(\pi)-  \hat J_{\alpha}(\hat \pi_\alpha)+    \hat J_{\alpha}(\hat \pi_\alpha)- J_{\alpha}(\hat \pi_\alpha)\\
    &\leq J_{\alpha}(\pi)-  \hat J_{\alpha}(\pi) +    \hat J_{\alpha}(\hat \pi_\alpha)- J_{\alpha}(\hat \pi_\alpha) \tag{Definition of $\hat \pi_\alpha$} \\
    &\stackrel{\text{(i)}}{\leq} J_{\alpha}(\pi)-  \hat J_{\alpha}(\pi). \tag{Pessimism}
\end{align*}
Here, in the step (i), we use 
\begin{align*}
  \forall x\in \Xcal_{\pre};  |r(x) - \hat r(x)| \le g(x). 
\end{align*}
Then, 
\begin{align*}
    J_{\alpha}(\pi)- J_{\alpha}(\hat \pi_\alpha)   & \leq J_\alpha(\pi) - \hat{J}_\alpha(\pi)  \leq 2\EE_{x\sim \pi} [\hat g(x)] \\
    &\leq 2 \sqrt{\EE_{x\sim \pi} [\{\hat g(x)\}^2]} \tag{Jensen's inequality} \\ 
    & \leq 2 \sqrt{ \left\|\frac{\pi}{p_{\pre} }\right\|_{\infty}\EE_{x\sim p_{\pre}}[\{\hat g(x)\}^2] }  \tag{Importance sampling}. 
\end{align*}
Hence, the statement is concluded.

\section{Theoretical Guarantees with Gaussian Processes} \label{sec:GPs}

We explain the theoretical guarantee when using Gaussian processes. In this section, we suppose the model is well-specified. 
\begin{assum}
   $ y = r(x) + \epsilon$ where $\epsilon\sim \Ncal(0,I)$ where $r$ belongs to an RKHS in $\Hcal_k$.  
\end{assum}

\subsection{Preparation}

We introduce the notation to state our guarantee. For details, see \citet[Appendix B]{srinivas2009gaussian}, \citet[Chapter 6.2]{uehara2021pessimistic}, \citet[Chapter C.3]{chang2021mitigating}.

For simplicity, we first suppose the following. 
\begin{assum}
   The space $\Xcal$ is compact, and $\forall x \in \Xcal; k(x,x)\leq 1$. 
\end{assum}

We introduce the following definition. Regarding details, refer to \citet[Chapter 12]{wainwright2019high}. 
\begin{definition}
   Let $\Hcal_k$ be the RKHS with the kernel $k(\cdot,\cdot)$. We denote the associated norm and inner product by $\|\cdot\|_k,\langle \cdot,\rangle_k$, respectively. We introduce analogous notations for 
   \begin{align*}
        \hat k(x,x')= k(x,x')-\kb(x)^{\top}\{ \Kb +\lambda I \}^{-1}  \kb(x'). 
   \end{align*}
   and denote the norm and inner product by $\|\cdot\|_{\hat k},\langle \cdot,\rangle_{\hat k}$. 
\end{definition}
Note as explained in  \citet[Appendix B]{srinivas2009gaussian} and \citet[Chapter C.3]{chang2021mitigating}, actually, we have $\Hcal_k =\Hcal_{\hat k}$.

In the following, We use the feature mapping associated with an RKHS $\Hcal_k$. To define this, from Mercer's theorem, note we can ensure the existence of orthonormal eigenfunctions and eigenvalues $\{\psi_i,\mu_i\}$ such that 
\begin{align*}
    k(\cdot,\diamond)=\sum_{i=1}^{\infty} \mu_i \psi_i(\cdot ) \psi_i(\diamond), \begin{cases} \int\psi_i(x)\psi_i(x)p_{\spp}(x)dx =1 \\ \int\psi_i(x)\psi_j(x)p_{\spp}(x)dx =0 (i\neq j)   \end{cases}. 
\end{align*}
Then, we define the feature mapping: 
\begin{definition}[Feature mapping]
    \begin{align*}
        \phi(x):=[\sqrt{\mu_1}\psi_1(x), \sqrt{\mu_1}\psi_1(x),\cdots]^{\top}. 
    \end{align*}
\end{definition}
Assuming eigenvalues are in non-increasing order, we can also define the effective dimension following \citet[Appendix B]{srinivas2009gaussian}, \citet[Chapter 6.2]{uehara2021pessimistic}, \citet[Chapter C.3]{chang2021mitigating}:  
\begin{definition}[Effective dimension]
\begin{align*}
    d' = \min_{j} \left \{j \in \mathbb{N}: j \geq  n\sum_{k=j}^{\infty}\mu_k\right \}. 
\end{align*}
\end{definition}
The effective dimension is commonly used and calculated in many kernels \citep{valko2013finite}. In finite-dimensional linear kernels $\{x\mapsto a^{\top}\phi(x):a\in \RR^d \}$ such that $k(x,z) =\phi^{\top}(x)\phi(z)$, letting 
$d'\coloneqq \rank(\EE_{x \sim p_{\spp} }[\phi(x)\phi(x)]$), we have 
\begin{align*}
     d' \leq \tilde d \leq d 
\end{align*}
because there exists $\mu_{\tilde d+1}=0,\mu_{\tilde d+2} =0,\cdots$.

\subsection{Calibrated oracle}

Using a result in \citet{srinivas2009gaussian}. we show 
\begin{align*}
        \hat r(x)-r(x) \leq C(\delta)\sqrt{\hat k(x,x)} 
\end{align*}
where
\begin{align*}
    C(\delta) = c_1 \sqrt{1+ \log^3(n/\delta) \Ical_{n}},\quad \Ical_n = \log(\det(I+\Kb)). 
\end{align*}
Then, with probability $1-\delta$, we have 
\begin{align*}
    \hat r(x)-r(x) & = \langle \hat r(\cdot)-r(\cdot), \hat k(\cdot,x)\rangle_{\hat k}  \tag{Reproducing property } \\ 
    & \leq   \|\hat r(\cdot)-r(\cdot)\|_{\hat k} \times \|\hat k(\cdot,x)\|_{\hat k}  \tag{CS inequality } \\
    &\leq \|\hat r(\cdot)-r(\cdot)\|_{\hat k}\sqrt{\hat k(x,x)} \\
    & \leq C(\delta) \sqrt{\hat k(x,x)}. \tag{Use Theorem 6 in \citet{srinivas2009gaussian}} 
\end{align*}

\subsection{Regret Guarantee (Proof of Corollary \pref{cor:GPS}) }

Recall from the proof of \pref{thm:key2}, 
\begin{align*}
    J_{\alpha}(\pi)- J_{\alpha}(\hat \pi_\alpha) \leq 2\EE_{x\sim \pi} [\hat g(x)]=2C(\delta) \EE_{x\sim \pi} [\sqrt{\hat k(x,x)} ]. 
\end{align*}

Now, first, to upper-bound $\EE_{x\sim \pi} [\sqrt{\hat k(x,x)} ]$, we borrow Theorem 25 in \citet{chang2021mitigating}, which shows 
\begin{align*}
    \EE_{x\sim \pi} [\sqrt{\hat k(x,x)} ]\leq c_1\sqrt{\frac{\tilde C_{\pi}d'\{d' + \log(c_2/\delta) \}}{n} }. 
\end{align*}
where 
\begin{align*}
\tilde C_{\pi}:=\sup_{\kappa:\|\kappa \|_2=1 }\frac{\kappa ^{\top}\EE_{x\sim \pi}[\phi(x) \phi^{\top}(x) ] \kappa}{\kappa ^{\top} \EE_{x\sim p_{\spp}}[\phi(x) \phi^{\top}(x)] \kappa}
\end{align*} 
Next, in order to upper-bound $C(\delta)$, we borrow Theorem 24 in \citet{chang2021mitigating}, which shows 
\begin{align*}
    \Ical_n \leq c_1\{d' + \log(c_2/\delta)\}d'\log(1+n). 
\end{align*}
The statement in Corollary~\ref{cor:GPS} is immediately concluded.

\section{Additional Details of Experiments} \label{sec:experiments}

\subsection{DNA/RNA sequences}

In this subsection, we add the details of experiments in \pref{sec:main_experiments}. 

\subsubsection{Architecture of Neural Networks}

\paragraph{Diffusion models.}

Regarding diffusion models for sequences, we adopt the architecture and algorithm tailored to biological sequences over the simple space \citep{avdeyev2023dirichlet}. Its architecture is described in Table~\ref{tab:diffusion}. 

\begin{table}[!h]
    \centering
        \caption{Basic architecture of networks for diffusion models}\label{tab:diffusion}
    \begin{tabular}{cccc}  \toprule 
    Layer   &  Input dimension &  Output dimension & Explanation  \\   \hline 
      1    &  $200 \times 4 $  &  $256$ &  Linear + ReLU    \\ 
      2    &  $256$  &  $256$   &  Conv1D + ReLU \\  
      $\cdots$ & $\cdots$ & $\cdots$ & $\cdots$   \\
      10   & $256$ & $256$ & Conv1D + ReLU   \\
      11 & $256$ & $256$ & ReLU  \\  \bottomrule 
    \end{tabular}
\end{table}

\paragraph{Oracles.} 

We use the architecture in \citet{avsec2021effective}, which is a state-of-the-art model in sequence modeling. We just change the last layer so that it is tailored to a regression problem as in \citet{lal2024reglm}.

\subsubsection{Hyperparameters}

In all experiments, we use A100 GPUs. The important hyperparameters are summarized in the following table (\pref{tab:important_hyper}).

\begin{table}[!h]
    \centering
     \caption{Important hyperparameters for fine-tuning. For all methods, we use Adam as an optimizer. }
    \label{tab:important_hyper}
    \begin{tabular}{c|c|c }  \toprule 
   Method  &   Type     &   Value  \\ 
   \hline
  \multirow{8}{*}{\alg}  &   Batch size  & $128$  \\
  & KL parameter $\alpha$  & $0.001$   \\ 
  & LCB parameter (bonus) $c$ & $0.1$ (UTRs), $0.1$ (Enhancers)     \\  
   & Number of bootstrap heads &  $3$\\  
   & Sampling to neural SDE &  Euler Maruyama \\ 
     & Step size (fine-tuning)  & $50$   \\ 
     \hline 
 \multirow{2}{*}{Guidance}   & Guidance level  &  $10$  \\ 
   & Guidance target & Top $5\%$  \\   \bottomrule 
    \end{tabular}
\end{table}

\paragraph{Hyperparameter selection.}

The process of selecting hyperparameters in offline RL is known to be a challenging task \citep{rigter2022rambo,paine2020hyperparameter} in general. A common practice in existing literature is determining crucial hyperparameters with a limited number of online interactions. In our case, the key hyperparameters include the strength of the LCB parameter (utilized online in \textbf{BRAID-Bonus}) and the termination criteria during training (applied to all fine-tuning algorithms such as \textbf{STRL}). To ensure a fair comparison, we operate within a framework where we can utilize $120 \times 20$ online samples. This implies that, for instance, in \textbf{STRL} and \textbf{BRAID-Bonus}, we conduct an online evaluation using $120$ samples for $20$ pre-defined epochs. However, in \textbf{BRAID-Bonus}, given the additional hyperparameters to be tested (strengths of the bonus term $0.01, 0.1, 1.0$), we use $40$ samples for each $20$ pre-defined epoch.

\subsubsection{Ablation Studies}

We performed ablation studies by varying the strength of the bonus parameter $C$ in \pref{fig:abletation_UTR}. We chose the one with the best performance in the main text (See the previous section to see the validity of this procedure). 

\begin{figure}[!h]
    \centering
    \includegraphics[width=0.8\textwidth]{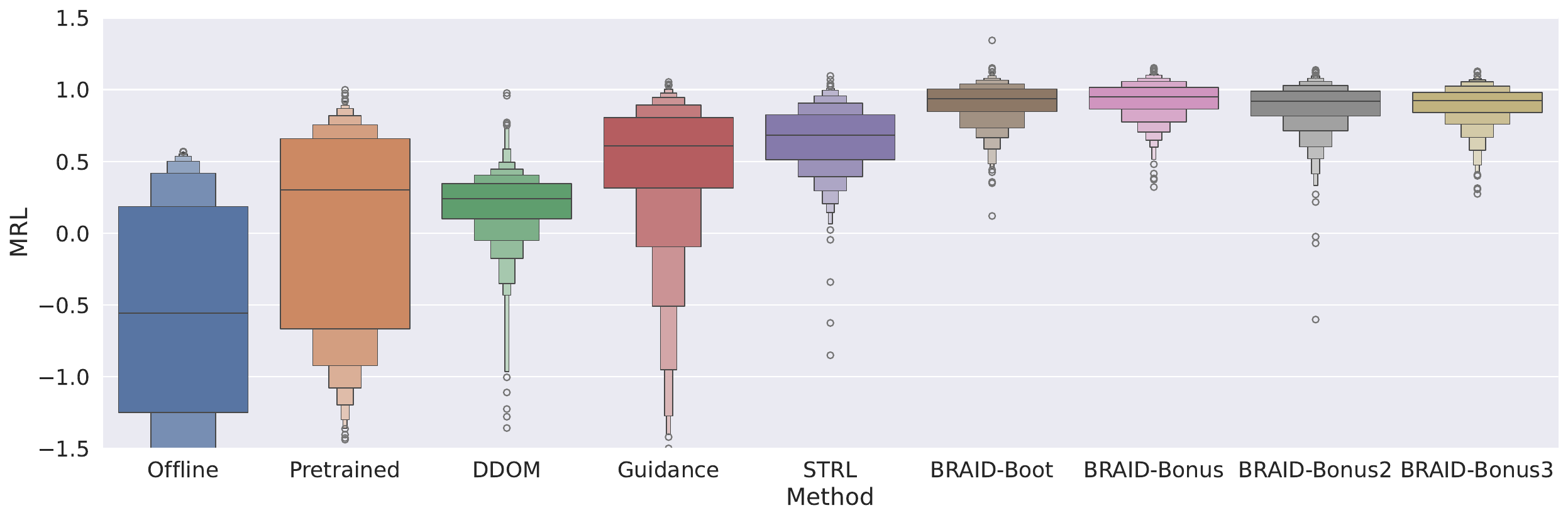}
    \caption{UTRs}
    \label{fig:abletation_UTR}
\end{figure}

\begin{figure}[!h]
    \centering
    \includegraphics[width=0.8\textwidth]{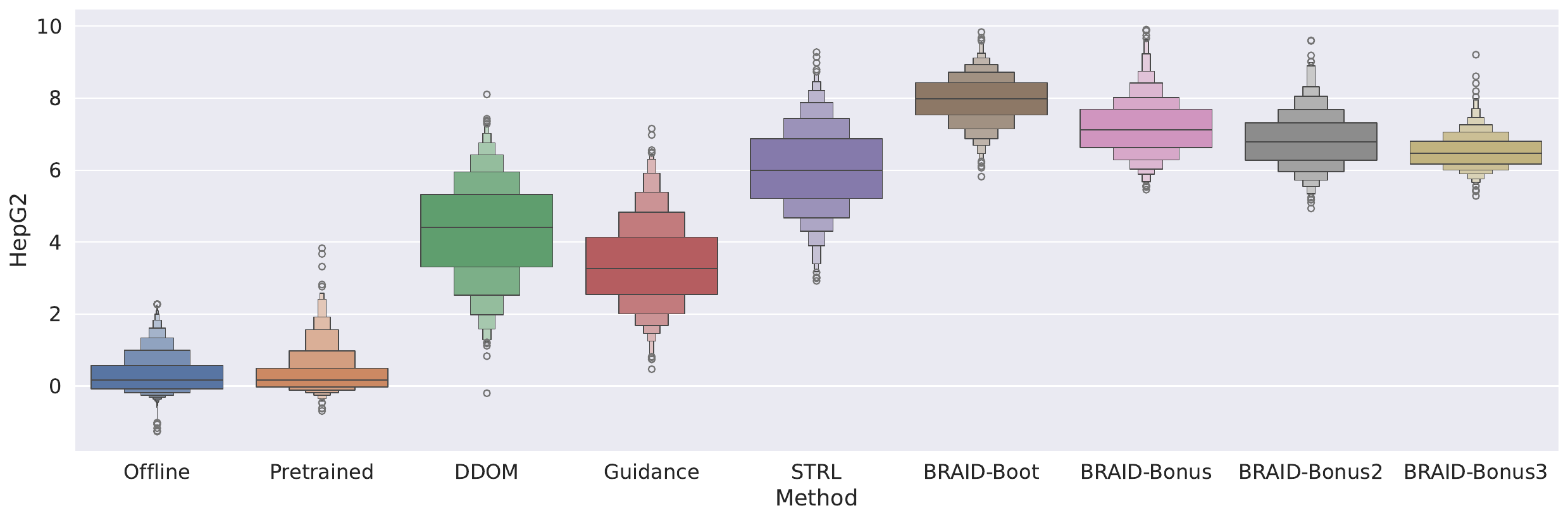}
    \caption{Enhancers}
    \label{fig:abletation_UTR}
\end{figure}

\subsection{Images}\label{sec:images_additional}

In this section, we describe the additional experiment regarding image generation in \pref{sec:images}.

\subsubsection{Description of Offline Data}

We utilize images from the AVA dataset~\citep{murray2012ava} as samples $x$, containing over $250,000$ image aesthetic evaluations. Rather than using the raw scores directly from the dataset, we derive the labels $y$ by utilizing the pre-trained LAION Aesthetic Predictor V2~\cite{schuhmann2022laion} built on top of CLIP embeddings. This choice is made because we employ the LAION Aesthetic Predictor as the ground truth scorer to assess both our methods and baselines. In total, we have curated an offline dataset comprising $255490$ image-score pairs: $\{x^{(i)},y^{(i)}\}$.

\subsubsection{Architecture of Neural Networks} 

We adopt the standard StableDiffusion v1.5~\citep{Rombach_2022_CVPR} as the pre-trained model with the DDIM scheduler~\citep{song2020denoising}. Note this pre-trained model is a conditional diffusion model. 

Using the offline dataset $\{(x^{(i)},y^{(i)})\}$, we train the reward oracle $\hat{r}$ by an MLP on the top of CLIP embeddings. The detailed MLP structure is listed in Table~\ref{tab:image_oracles}. Note that, compared to the true LAION Aesthetic Score Predictor V2~\citep{schuhmann2022laion}, our reward oracle proxy has a simpler structure with fewer hidden dimensions and fewer layers. We aim to impose the hardness of fitting the true reward model, which is typically infeasible in many applications. In such scenarios, a pessimistic reward oracle is especially beneficial to mitigate overoptimization.

\begin{table}[!h]
    \centering
    \caption{Architecture of reward oracle for aesthetic scores}\label{tab:image_oracles}
    \begin{tabular}{cccc}
    \toprule
 Layer   &  Input dimension &  Output dimension & Explanation  \\   \hline 
  1    &  $768 $  &  $256$ &  Linear + ReLU    \\
  2    &  $256 $  &  $64$ &  Linear + ReLU    \\
  3    &  $64 $  &  $16$ &  Linear + ReLU    \\
    4    &  $16$  &  $1$ &  Linear     \\
    \bottomrule
    \end{tabular}
\end{table}

\subsubsection{LLM-aided evaluation}\label{subsec:LLAVA_guided}

As stated in the main text, the original LAION Aesthetic Predictor V2~\citep{schuhmann2022laion} tends to assign higher scores even to images that disregard the original prompts, which is undesirable. To effectively identify such problematic scenarios, we employ a pre-trained multi-modal language model to verify whether the original prompt is present in the image or not. For each generated image, we provide the following prompt to LLaVA~\citep{liu2024visual} along with the image:
\begin{verbatim}
<image>
USER: Does this image include {prompt}? Answer with Yes or No
ASSISTANT:
\end{verbatim}
We evaluated its accuracy and precision with human evaluators by generating images using Stable Diffusion with animal prompts (such as dog or cat). The achieved F1 score was $1.0$.

\subsubsection{Hyperparameters}\label{subsec:hyper}

In all image experiments, we use four A100 GPUs for fine-tuning StableDiffusion v1.5~\citep{Rombach_2022_CVPR}. The set of training hyperparameters is listed in Table~\ref{tab:hyper_params}.

\begin{table}[!h]
    \centering
     \caption{Important hyperparameters for fine-tuning Aesthetic Scores. }
    \label{tab:hyper_params}
    \begin{tabular}{c|c|cc}\toprule 
      Method     &  Parameters     & Values \\ \midrule  
  \multirow{6}{*}{\textbf{\alg}} 
  & Guidance weight & $7.5$ \\
  & DDIM Steps  & $50$  \\ 
  &  Batch size  & $128$   \\
  & KL parameter $\alpha$  & $1$   \\ 
  & LCB bonus parameter $C$ & $0.001$  \\
   & Number of bootstrap heads & $4$ \\    \midrule 
\multirow{4}{*}{ \textbf{STRL}}  
 & Guidance weight & $7.5$ \\  
  & DDIM Steps & $50$\\
& Batch size  & $128$   \\
 & KL parameter $\alpha$ & 1\\
  \midrule  
 \multirow{2}{*}{\textbf{Offline guidance}}
 & Guidance level  &  $100$ \\
  & Guidance target  &  $10$\\
  \midrule  
\multirow{6}{*}{ \textbf{Optimization} } & Optimizer  & AdamW \\
 & Learning rate  & $0.001$ \\
 & $(\epsilon_1, \epsilon_2)$ &  $(0.9, 0.999)$\\ 
 & Weight decay & $0.1$\\
 & Clip grad norm & $5$\\
 & Truncated back-propagation step $K$ & $K \sim \text{Uniform}(0,50)$\\
  \bottomrule  
    \end{tabular}
\end{table}

\subsubsection{Effectiveness of LLaVA-aided evaluation} 
 
In our evaluation, we utilize a large multi-modal model like LLaVA. As previously mentioned, relying solely on the raw score fails to detect scenarios where generated images ignore the given prompts.

\begin{figure}[!h]
    \centering
\includegraphics[width=0.8\textwidth]{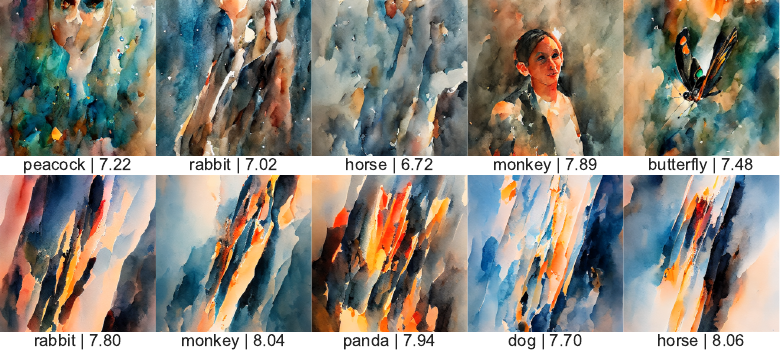}
    \caption{Image-prompt alignment failures detected by LLaVA.}
    \label{fig:llava-failures}
\end{figure}

Table~\ref{tab:llava_eval} illustrates the outcomes of LLaVA-assisted evaluations for the pre-trained model and four checkpoints of the \textbf{STRL} baseline. It is evident that LLaVA successfully identifies all samples generated by the pre-trained model and the first two checkpoints. However, despite seemingly high-reward samples, many samples from checkpoints 3 and 4 do not align correctly with their prompts, resulting in a reduced mean reward. Figure~\ref{fig:llava-failures} showcases five failure examples from each of checkpoints 3 and 4. Thus, we can validate our quantitative evaluation of reward overoptimization.

\begin{table}[!h]
    \centering
    \caption{Statistics of LLaVA-adjusted scores.}\label{tab:llava_eval}
    \begin{tabular}{ccccc}
    \toprule
 method   &  mean &  min & max & invalid/total samples  \\   \hline 
  pre-trained model    &  $5.789 $  &  $4.666$ &  $6.990$ & $0/400$\\ 
  STRL-ckpt-1    &  $6.228 $  &  $4.769$ & $7.193$  & $0/400$\\
   STRL-ckpt-2    &  $6.870 $  &  $5.892$ &  $7.602$ & $0/400$\\
   STRL-ckpt-3    &  $6.484 $  &  $0.0$ &  $7.944$ & $50/400$\\
   STRL-ckpt-4    &  $0.200 $  &  $0.0$ &  $7.620$ & $389/400$\\
    \bottomrule
    \end{tabular}
\end{table}

\subsubsection{Ablation Studies}\label{subsec:abelation}

\paragraph{Ablation on \alg-Bonus hyperparameter}

\begin{figure}[!h]
    \centering
    \includegraphics[width=0.9\textwidth]{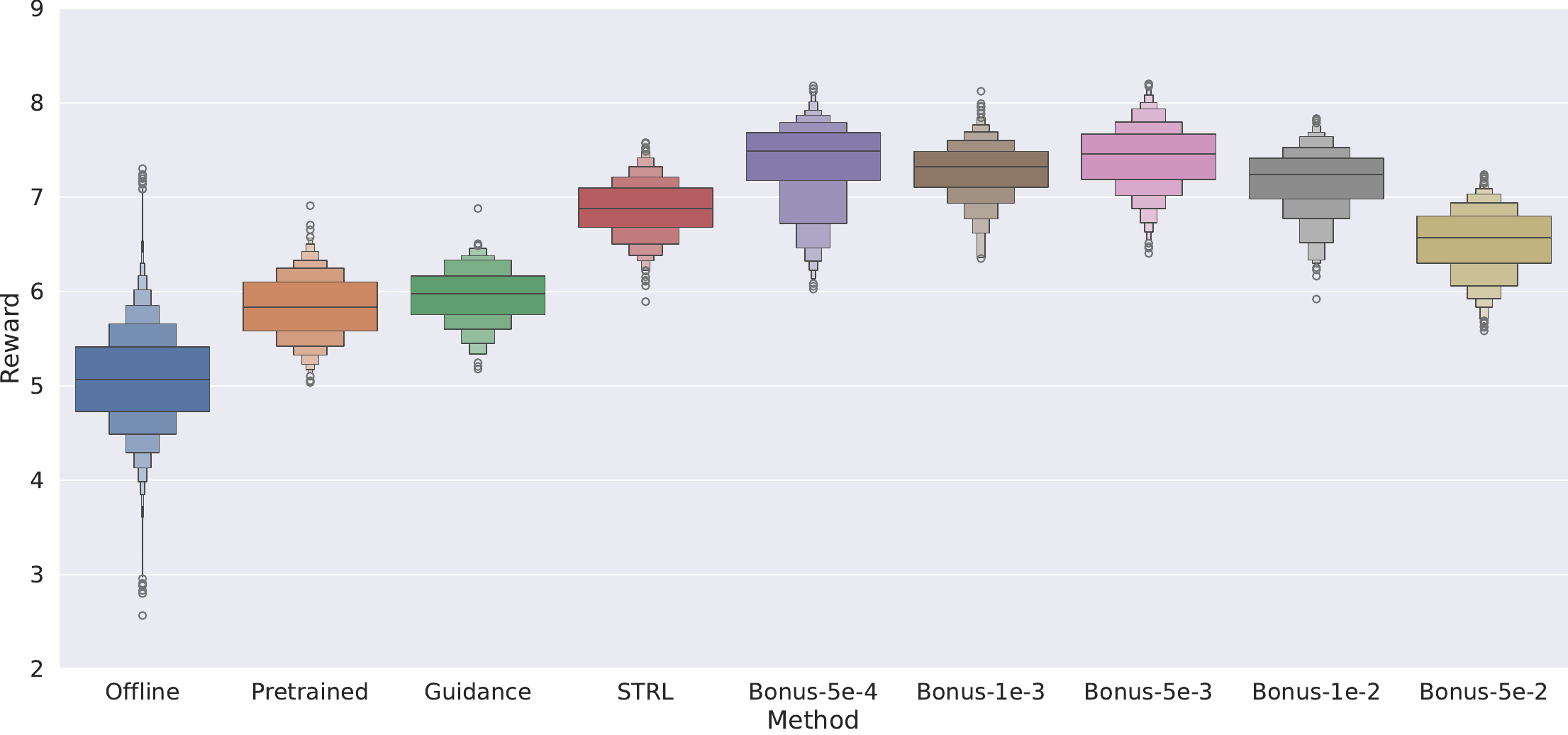}
    \caption{Ablation study for \textbf{\alg-Bonus}. By adjusting the pessimism strength $C_1$ while keeping $\lambda=0.1$, we show that \textbf{\alg-Bonus} outperforms all baselines for a wide range of hyperparameter selection.}
    \label{fig:ablation}
\end{figure}

We provide the boxplots for different Bonus hyperparameters in Figure~\ref{fig:ablation}, indicating our method's robustness to hyperparameter tuning.

\begin{figure}[!h]
    \centering
    \includegraphics[width=0.9\textwidth]{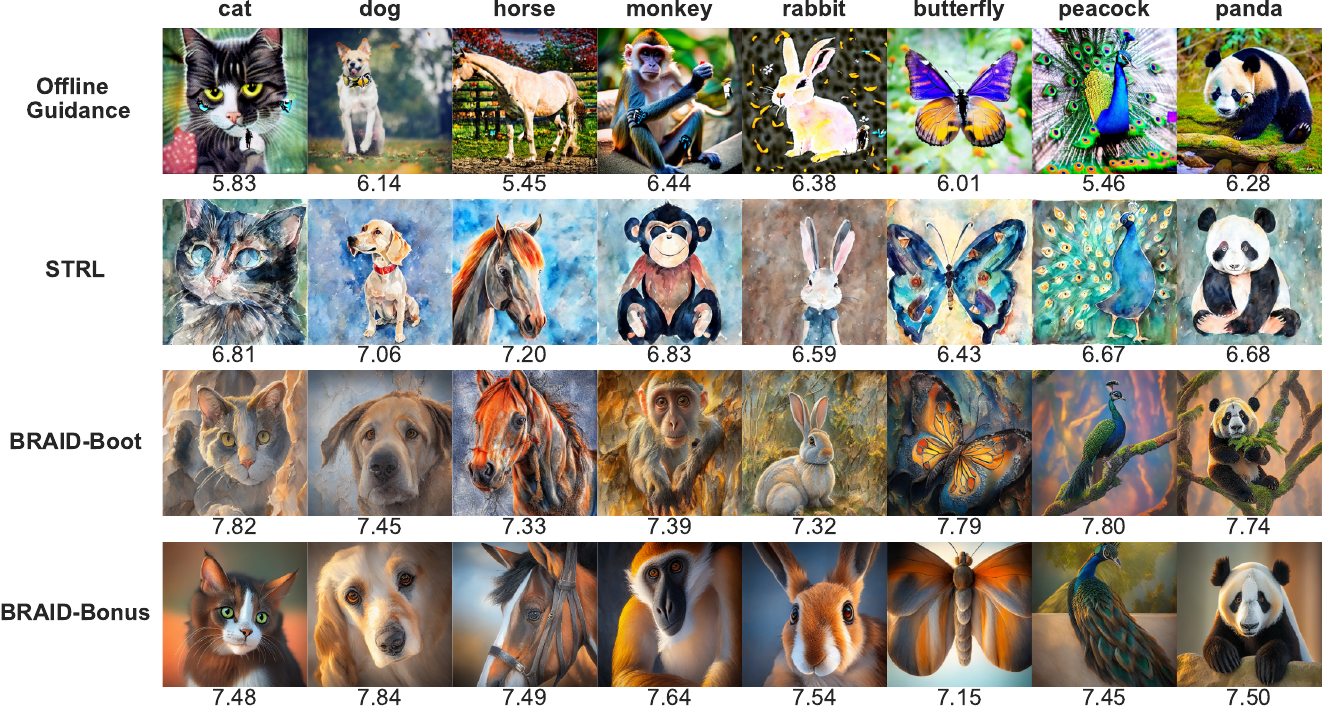}
    \caption{More images generated by \textbf{\alg}~and baselines. All algorithms choose the best checkpoint according to our LLaVA-aided evaluation. The visualization demonstrates the benefits of introducing pessimistic terms that can help to achieve high scores while mitigating reward overoptimization.}
    \label{fig:more_vis}
\end{figure}

\paragraph{Additional qualitative results}
More image visualizations for \textbf{\alg}~and baselines can be found in Figure~\ref{fig:more_vis}.

\end{document}